\documentclass[10pt,twocolumn,letterpaper]{article}

\usepackage{iccv}
\usepackage{multirow}
\usepackage{times}
\usepackage{graphicx}
\usepackage{amsmath}
\usepackage{amssymb}
\usepackage{algorithmic}
\usepackage{textcomp}
\usepackage{xcolor}
\usepackage{colortbl}
\usepackage{enumitem}
\usepackage{media9}

\usepackage[ruled,vlined]{algorithm2e}

\SetCommentSty{mycommfont}
\SetKwInput{KwInput}{Input}                % Set the Input
\SetKwInput{KwOutput}{Output}              % set the Output

\usepackage[pagebackref=true,breaklinks=true,letterpaper=true,colorlinks,bookmarks=false]{hyperref}

\iccvfinalcopy % *** Uncomment this line for the final submission

\def\iccvPaperID{5973} % *** Enter the ICCV Paper ID here

\definecolor{applegreen}{rgb}{0.55, 0.71, 0.0}
\definecolor{candyapplered}{rgb}{1.0, 0.03, 0.0}
\definecolor{limegreen}{rgb}{0.2, 0.8, 0.2}
\definecolor{melon}{rgb}{0.99, 0.74, 0.71}
\usepackage{color}

\ificcvfinal\fi
\begin{document}

% For a paper whose authors are all at the same institution,
% omit the following lines up until the closing ``}''.
% Additional authors and addresses can be added with ``\and'',
% just like the second author.
% To save space, use either the email address or home page, not both

%%%%%%%%% TITLE
\title{Counting and Segmenting Sorghum Heads}

\author{
Min-hwan Oh\\
Columbia University\\
New York, NY USA\\
{\tt\small m.oh@columbia.edu}
\and
Peder Olsen\\
IBM Research\\
Yorktown Heights, NY USA\\
{\tt\small pederao@gmail.com}
\and
Karthikeyan Natesan Ramamurthy\\
IBM Research\\
Yorktown Heights, NY USA\\
{\tt\small knatesa@us.ibm.com}
}

\maketitle
%\thispagestyle{empty}

%%%%%%%%% ABSTRACT
\begin{abstract}
    Phenotyping is the process of measuring an organism's observable traits. Manual phenotyping of crops is a labor-intensive, time-consuming, costly, and error prone process. Accurate, automated, high-throughput phenotyping can relieve a huge burden in the crop breeding pipeline. In this paper, we propose a scalable, high-throughput approach to automatically count and segment panicles (heads), a key phenotype, from aerial sorghum crop imagery. Our counting approach uses the image density map obtained from dot or region annotation as the target with a novel deep convolutional neural network architecture. We also propose a novel instance segmentation algorithm using the estimated density map, to identify the individual panicles in the presence of occlusion. With real Sorghum aerial images, we obtain a mean absolute error (MAE) of 1.06 for counting which is better than using well-known crowd counting approaches such as CCNN, MCNN and CSRNet models. The instance segmentation model also produces respectable results which will be ultimately useful in reducing the manual annotation workload for future data.
    
    % Detection of panicle regions in the presence of occlusion, as well as counting is performed using the image density map.
\end{abstract}

%%%%%%%%% BODY TEXT
\section{Introduction}

Genotyping and phenotyping constitute two key components of plant breeding. Genotyping involves the understanding of the genetic constitution of plants, whereas phenotyping involves the measurement of their observable traits.  While genotyping has become more accurate and affordable, phenotyping has become the bottleneck in accelerated breeding programs \cite{FURBANK2011635}.  This is because manual phenotyping methods are labor-intensive, inaccurate, and expensive.

In this paper, we develop methods to automatically count and segment panicles for Sorghum crops using a novel two-stage convolutional neural network (CNN) architecture that uses the panicle density map as target. Panicles are the heads of the plant that carry the grain. They are one of the most important phenotypes in crop breeding, since they correlate highly to the grain yield \cite{maman2004yield}, which is directly useful for food
and feed. Panicles also constitute a substantial fraction of the biomass, which can be used for sustainable bio-fuel production. The work presented here is incorporated in a high-throughput phenotyping pipeline being developed as a part of the Department of Energy's ``Transportation Energy Resources from Renewable Agriculture Phenotyping'' (TERRA) program, which also funded our work and collaboration with Purdue University.

\begin{figure}[h]
  \begin{center}
  \includegraphics[width=\linewidth]{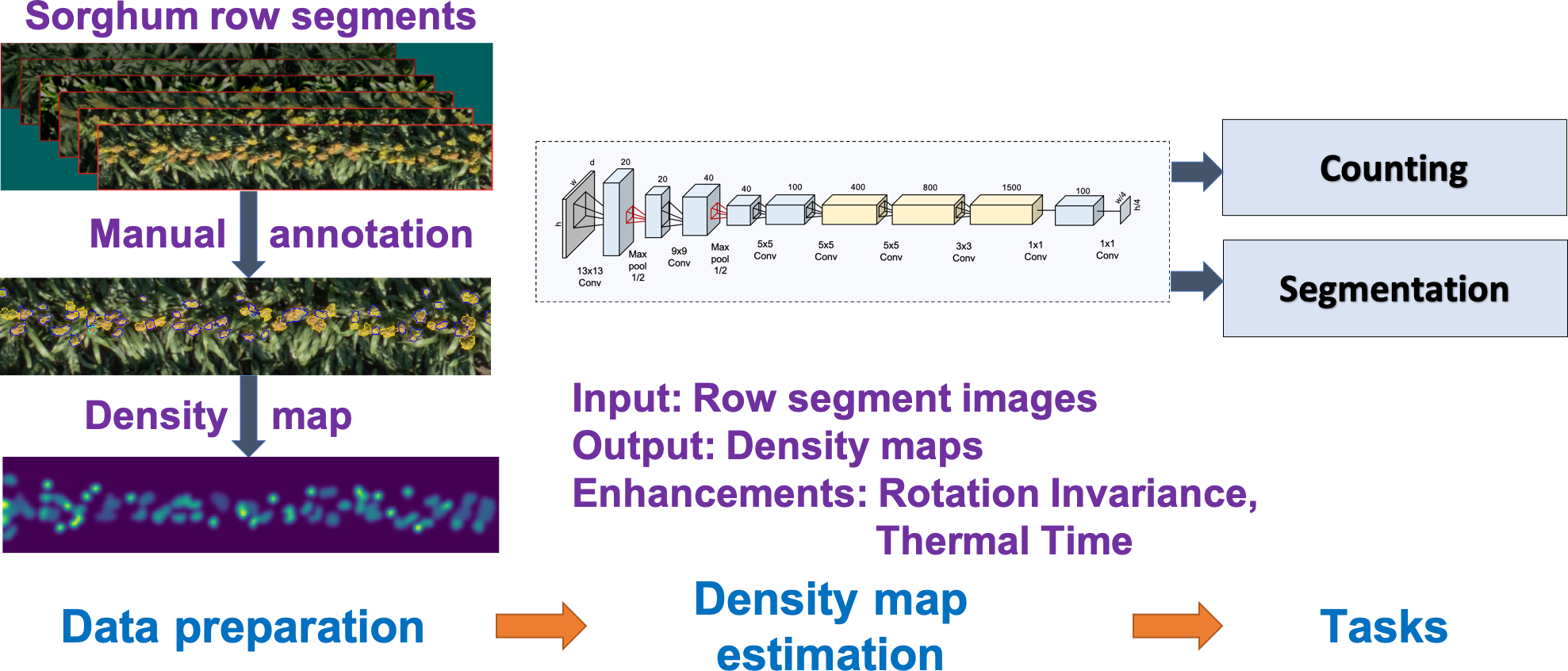}
  \end{center}
  \caption{Overview of our proposed approach for counting and segmenting panicles of Sorghum.}\label{fig:overview}
\end{figure}

\subsection{Overview and Contributions}
An overview of our system is provided in Figure \ref{fig:overview}. We first create superpixel sets of various sizes from the images to do human annotation to create the training data set\footnote{Video of our panicle pixel annotation tool for panicle pixel detection: \url{https://youtu.be/McMRqPDyQjE} and for instance segmentation: \url{https://youtu.be/B6wxXUfrUuw}.}. This annotation can be either dot-based or region-based. We create image density maps for counting using these annotations, and train a novel two-stage CNN architecture to predict these density maps. The proposed architecture includes methods to impose prior knowledge such as monotonicity in counts, and temperature dependence on the growth of the crop. The first stage is a panicle pixel detector that feeds into the second density estimator stage. Counts are obtained by integrating the density maps. Our proposed counting approach outperforms state-of-the-art systems such as CCNN \cite{Onoro:16}, MCNN \cite{zhang2016single}, and CSRNet \cite{li2018csrnet} that are primarily applied to crowd counting. We then segment the individual panicle instances using a novel occlusion-aware clustering approach with the estimated density maps and detected panicle regions as inputs. Hence, the two outputs of our system is image level counts and the segmented individual panicle instances. The panicle instances can then be fed back to the annotation tool to accelerate human annotation. We consider our focus on Sorghum panicles to be an important contribution in this work, since it has the potential to directly impact crop breeding.

%However, we note that the segmentation system is not extremely accurate and is not used for phenotyping.   

%We choose to train a regression network first using the image density map as target, and then segment panicles based on this output. To the best of our knowledge, our approach is the first \textit{segmentation by regression} method. %\karthi{One could also choose to segment the panicles first and then count them, but there are two key impediments here: (a) Significant variation in the panicle images across varieties, which makes segmentation hard, (b) considerable  occlusion among panicles, which will result in an underestimate of the count.}

Some of the key challenges in counting and segmenting aerial Sorghum imagery include: (a) Dramatic differences in appearance between panicles. The appearance varies in size ($20\mathrm{cm}^2$ to $1000\mathrm{cm}^2$), shape (spindle, broom, cylinder or a lax cone), color (chalky white, green, yellow, rusty brown or black), pose, and grain-size, (b) self-occlusion of panicles later in the season, which is particularly severe with grassy varieties (see Appendix \ref{sec:example_images}). Hence, to develop a good panicle counting and segmentation system, we require a diverse, accurately annotated data set, and robust machine learning models trained on this data set, which is the main focus of this paper.

We obtain a mean absolute error (MAE) of $1.06$ compared to $1.19$ for the performance of a CSRNet model for counting panicles from aerial images of Sorghum. The instance segmentation model produces an mAP of $0.66$ at an IoU detection threshold of $0.5$.  This nearly free instance segmentation can ultimately reduce the manual annotation workload for future data collection, by integration into the manual annotation tool.

We also obtained publicly available aerial imagery of Sorghum used for panicle counting \cite{guo2018aerial}. Results with this data show that the deep counting methods, including the proposed counting regressor, outperform the methods proposed in \cite{guo2018aerial} in a vast majority of cases.

% \noindent \textbf{Contributions:} Our main contributions in this paper are listed below. \begin{itemize}[noitemsep]
%     \item 
%     We propose a new CNN-based approach to learn image densities from dot- or region-based annotations.  This system takes as input the RGB channels, an added thermal channel and a predicted panicle detection map (from another CNN system).
%     The image densities can then be directly used for counting.  The proposed CNN architecture includes methods to impose monotonicity in counts, and temperature dependence on the growth of the crop.
%     \item The inferred density maps are also used to obtain panicle instance segmentation, in a highly occluded regime, using a novel density aware greedy clustering algorithm, using the proposed \textit{segmentation by regression} method.
%     %\item We demonstrate high performances for the panicle counting using aerial images of several varieties of the Sorghum crop.
% \end{itemize}

In Section \ref{sec:rel_work}, we discuss the related work in counting and instance segmentation. Section \ref{sec:data} provides detailed description of the data and the steps involved in preparing it for modeling. Section \ref{sec:cnn_arch} discusses the proposed approaches for counting along with counting experiments, while the proposed segmentation algorithm and the demonstration of its results is provided in Section \ref{sec:detect_segment}. 

\begin{figure*}[h]
  \begin{center}
  \includegraphics[width=\linewidth]{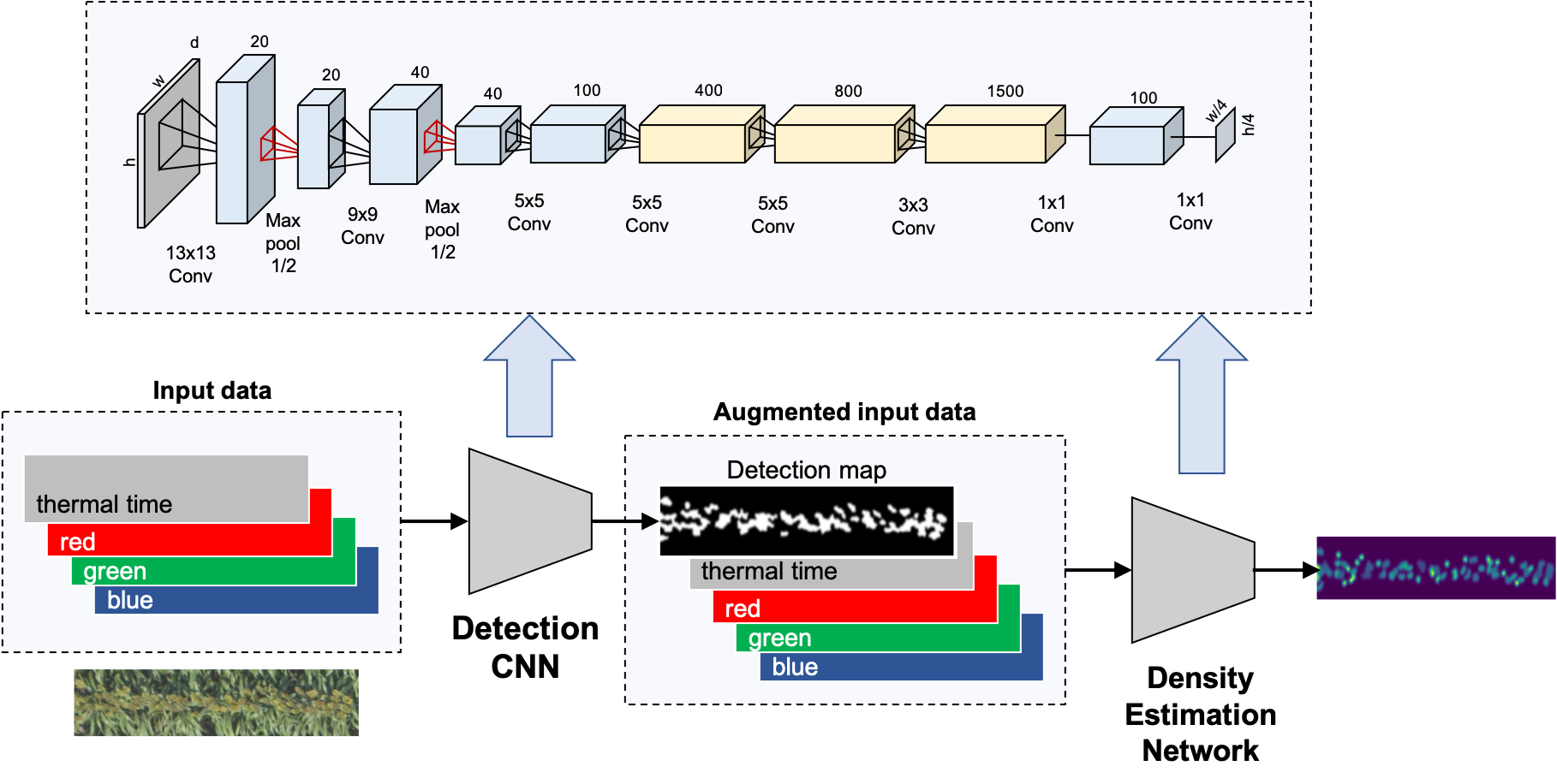}
  \end{center}
  \caption{Proposed two-stage CNN architecture for density estimation for counting and segmentation of panicles.}\label{fig:system}
\end{figure*}

% provides Our main contributions in this paper are

%\karthi{Like Peder mentioned, these problems could be circumvented if we simultaneously have 2 networks feeding into each other - one for segmentation, one for counting.}

\section{Related Work}
\label{sec:rel_work}

The problem of counting objects in images is seeing a resurgence with the advent of deep learning with applications in crowd and traffic monitoring, medical image analysis, and agronomy. The common challenge here is that manual counting can be laborious, inaccurate, and expensive.

{\em Counting by detection} is a 2-step process that first detects the objects then counts them. 
%This works very well for simple detection task where there is little or no occlusion and detectors are accurate. 
{\em Counting by regression} uses regression models, such as neural networks, with the object count as a target for the loss function \cite{Chan:12}. In {\em Counting by segmentation} the image is segmented into foreground and background, and the counts are estimated from the foreground \cite{hernandez2018using}. Recently proposed {\em Detection by regression} approaches \cite{xie2015beyond, xie2018microscopy} attempt to reconstruct an image density map \cite{Lempitsky:10} as well as detect the cell in microscopy images. 
%The image density map was first introduced in \cite{Lempitsky:10}, in a non deep learning setting. 
This is the closest to our proposed segmentation approach, however, our problem is much more complex since panicles are much more heterogeneous within and across the various varieties of Sorghum compared to cells that are mostly homogeneous in shape and appearance.

Most prevailing methods for counting use {\em counting by regression} approaches, where the non-linear regression function is represented by a CNN and the target is an image density map or the final count itself. However, some applications could use {\em counting by detection} or {\em counting by segmentation} if the detection or segmentation can be performed accurately. Depending on the problem, there is variation in the steps of the methodology such as estimating the density map from an annotation, the actual CNN architecture, and other enhancements involving semantic understanding of the counting problem. Some example application areas are in,
\begin{enumerate}

    \item{\bf Crowd and Traffic Monitoring}: Counting people in a crowd \cite{sindagi2018survey, zhang2016single, shang2016end, boominathan2016crowdnet} is perhaps the focus area of new counting approaches in many mainstream computer vision venues, and hence expanded in Section \ref{sec:crowd_count}. There is also a good amount of literature on methods used for counting vehicles \cite{chauhan2019embedded, tayara2018vehicle, chung2018image, guerrero2015extremely}.
    
    \item {\bf Medical imaging}: For general cell counting \cite{xue2016cell}, cell counting and detection using fully convolutional regression networks \cite{xie2018microscopy}, counting cells for images showing a developing human embryo \cite{Khan:16}, and counting bacteria colonies \cite{ferrari2015bacterial}.

    \item {\bf Agronomy}: For general plant counting \cite{Ribera:17}, counting palm trees \cite{Li:16}, plant stalks \cite{Baweja:18}, fruits \cite{Chen:17, Rahnemoonfar:17} and maize tassels \cite{Lu:17}. This is a less mainstream application for counting, but a rich domain, that has real impact in people's lives, and this will be the focus of our paper.
\end{enumerate}

For agronomy applications, such as ours, understanding the crop is fundamental to developing an effective algorithm. 
%We follow this principle in the development of our algorithm, and we believe this can be extended to other crops also. 
A well executed \textit{counting by detection} approach that illustrates this is \cite{Yamamoto:14} that count tomatoes by first segmenting the image then a decision tree extracts the fruit segments. The decision tree used color, shape, texture and size features to locate the tomatoes.  A similar approach was taken in \cite{guo2018aerial} for counting panicles in Sorghum.  This is the only other published work we are aware of for panicle counting.

There is also a huge body of literature in instance segmentation, and we will discuss a few important and well-known ones here. A notable work is Mask R-CNN \cite{he2017mask} proposed by  He \textit{et al.}, which extends the well-known region proposal network, Faster R-CNN \cite{ren2015faster}, by including a branch for predicting segmentation mask on each region of interest. In \cite{kulikov2018instance}, the authors reduce the problem of instance segmentation to semantic segmentation to leverage the rich works in that area, by assigning \textit{colors} to object instances. The information propagation in proposal based instance segmentation is boosted in \cite{Liu_2018_CVPR} by enhancing the feature hierarchy with localization signals. An instance level segmentation approach for video is proposed in \cite{NIPS2017_6636} that uses a recurrent neural net to take advantage of long-term temporal structures. An instance segmentation approach for neuronal cells in the brain using a hierarchical neural network was proposed in \cite{yi2018pixel}. A refreshing mathematical approach for instance segmentation using semi-convolutional operators was proposed in \cite{novotny2018semi}. To mitigate the labeling costs involved in instance segmentation, Hu \textit{et al.} \cite{hu2018learning} propose a partially supervised paradigm to learn from a large set of categories that have box annotations where only a small fraction of them have mask annotations. 

\subsection{Crowd Counting}
\label{sec:crowd_count}
We review several architectures for crowd counting, since it is the main application area for counting in mainstream computer vision conferences. In this work, we reuse some of these architectures for comparisons.

In crowd counting, a key challenge is to build effective CNN architectures that handle perspective scale distortions. The central building block for several counting systems is the counting CNN (CCNN) \cite{Onoro:16} that we also use and fine-tune in our work. The CCNN, shown in Figure \ref{fig:ccnn_arch}, is simply a deep CNN with an image density map as a regressor. A family of models have been developed that take an ensemble approach to the scale distortion problem. For the models in the family, the two main decisions that need to be made are: how to design the components in the architecture for the different head sizes, and how to combine these components.  

A simple but elegant member of this family is the multi-column CNN (MCNN) \cite{zhang2016single}, as seen in Figure \ref{fig:mcnn_arch}, where the component models in the ensemble are CCNNs with the receptive fields in the convolutional kernel designed for a particular head size. The resulting CCNN predictors are then combined into one density map with a fully convolutional layer. The CCNN tends to over estimate the count when the density is very low and under estimate the count near the horizon line where the density is extremely high. This was used advantageously in the DecideNet model \cite{liu2018decidenet} where they noted that the counting by detection approach is particularly good when the density is very low. They then used another neural network to estimate an attention map and combine the two approaches.

%This paper also extended the popular Shanghai crowd estimation data set, first introduced in \cite{Zhang:15}, to 1198 images containing 330,000 annotated heads.  

While these papers used the CCNN type architecture, the popular U-net architecture has been used in \cite{shen2018crowd} together with a scale consistency regularizer that forces collaborative predictions from the ensemble models. The U-net model has been widely used in medical imaging. \textit{E.g.},  Ronnenberger \textit{et al.}, \cite{ronneberger2015u} use the U-net for cell tracking in bio-medical imaging. Examples of works that allow for very deep networks include \cite{shi2018crowd} and \cite{li2018csrnet}, both of which use a VGG \cite{simonyan2014very} type architecture. Shi \textit{et al.} \cite{shi2018crowd} utilize a multi-scale architecture, where a part of their network uses complex ensemble averaging to combine image density maps from models built for different scales. Li \textit{et al.} \cite{li2018csrnet} introduce the dilated convolution to aggregate multi-scale contextual information in the CSRNet model (see Figure \ref{fig:CSR_arch}). \cite{oh2019crowd} extended CSRNet by adding parallel branches for the last predictive layer.  This allowed for uncertainty estimation as well as a way to improve the performance through ensemble averaging.  Finally, it is worth mentioning the approach taken in \cite{amirgholipour2018ccnn} that first estimates the head size and position in order to choose the hyper parameters for the CCNN.   

We have used the CCNN, MCNN and CSRNet models in our experiments, but it should be noted that none of these models achieve the best performance on the common crowd counting benchmarks.  The CSRNet was the best performing model when published in CVPR 2018, but has since been surpassed on several benchmarks by SANet \cite{wu2018adaptive}, ic-CNN \cite{ranjan2018iterative} and by ASD \cite{cao2018scale}.    Table~\ref{table:crowd} shows the performance of these 6 systems on the large and popular \texttt{UCF CC 50} benchmark.  The code for CCNN, MCNN and CSRNet have been released, but we are unaware of codes for SANet, ic-CNN, and ASD, which makes it almost impossible for us to make comparisons.  We believe CSRNet is the best system for crowd counting that have been independently verified.

\begin{table}[h]
\newcolumntype{x}{>{\columncolor{limegreen!30}}l}
\newcolumntype{s}{>{\columncolor{limegreen!60}}l}
\newcolumntype{y}{>{\columncolor{melon!30}}c}
\newcolumntype{t}{>{\columncolor{melon!60}}c}
\newcolumntype{z}{>{\columncolor{blue!20}}c}
\newcolumntype{u}{>{\columncolor{blue!40}}c}
\begin{center}
\begin{tabular}{|xy||xy|}
\hline
  \bf System & MAE & \bf System & MAE\\ \hline
CCNN & 488.7 & ic-CNN & 260.9 \\
MCNN & 377.6 & SANet & 258.4 \\
CSRNet & 266.1 & ASD & 196.2 \\ \hline
\end{tabular}
\end{center}
  \caption{Performance for crowd counting for 6 systems on the \texttt{UCF CC 50} benchmark.
  }\label{table:crowd}
\end{table}

All of these methods handle the problem of perspective by creating specialized models for heads of varying sizes. This will however degrade performance when all the human heads have the same size. In our application, the size of panicles varies through the season and by the variety, but we did not see any improvement in counting performance using the above approaches that compensate for perspective.

\begin{figure}[h]
  \begin{center}
  \includegraphics[width=\linewidth]{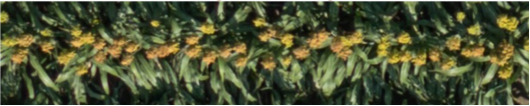}
  \end{center}
  \caption{An example row-segment image.}\label{fig:row_segment}
\end{figure}

\section{Data and Preparation}
\label{sec:data}
In this section, we describe the RGB image data of Sorghum crops collected using unmanned aerial vehicles (UAVs) in 2017 at the experimental fields of Purdue University. These experiments are a part of our collaborative project in high-throughput phenotyping. This is the data using which we demonstrate our proposed approaches for counting and segmentation. We also describe our data preparation methodology.  More details on the data are available in Appendix \ref{sec:example_images}.

More than 1000 varieties of sorghum were planted on the fields. We restrict our focus to a collection of plots named the  hybrid calibration panel where 18 commercial Sorghum varieties were planted (see Appendix \ref{sec:pedigrees}). These 18 varieties were chosen to represent the variations from across the larger set. The hybrid calibration panel consisted of 88 rows that were divided into 20 ranges, and each image covered one row-segment corresponding to a particular row and range.  Each of the 18 different varieties were planted on 4 plots, and each plot contained 12 consecutive row-segments on the same range. The 12 row-segments within each plot will be conveniently referred to as field rows, so that the first row-segment in a plot is field-row one, the second field-row two and so on. These row segments were extracted as described in \cite{Ribera:17}. As the panicle is the head of the plant it sits above the canopy and is visible by the UAV camera (See Figure \ref{fig:row_segment}).

%(\karthi{show one raw row segment image  without any annotation.})

%The four plots were not adjacent and were located on different ranges.  Moreover, these plots were all located in the interior as the outer ranges and four outer rows contained a protective layer of sorghum plants that was not used for phenotyping.

We used images collected on $6$ different dates collected by UAVs that are roughly 1 week apart (7/11, 7/17, 7/25, 8/2, 8/8, and 8/16), and covered the full panicle development stage for most varieties of Sorghum. We used field rows 2 and 3 of each plot for these six dates (7/11-8/16) where no destructive phenotyping was performed. These $2 \text{ rows} \times 4 \text{ plots} \times 6 \text{ dates} \times 18 \text{ varieties}$, a total of $864$ row-segments were used in our panicle counting experiments. We also augmented the training data by considering rotations by 90, 180, 270 and 360 degrees as well as reflections for each rotation. This increases the size of the data set eight-fold, resulting in a total of $6,912$. An alternative to data augmentation is to use group equivariant networks \cite{cohen2016group, cohen2018spherical}, but we stick to the simpler approach here.

All our experiments are performed using four way cross-validation, where we held out one plot for each variety for testing and trained on the rest, then rotated the held out plot until the entire data set was covered in the four test sets. The size of training and test sets for each cross-validation run would be $5,184$, and $1,728$ respectively. Finally, each row segment will have a unique result.
% See Figure~\ref{fig:data} for an example. 

The main challenges with counting using the data are: (a) Dramatic differences in appearance between panicles.  The appearance varies in size ($20\mathrm{cm}^2$ to $1000\mathrm{cm}^2$), shape (spindle, broom, cylinder or a lax cone), color (chalky white, green, yellow, rusty brown or black), pose, and grain-size, (b) self-occlusion of panicles later in the season, which is particularly severe with grassy varieties (see Appendix  \ref{sec:example_images}). Hence, to develop a good panicle counting and segmentation system, we require a diverse, accurately annotated data set, and robust machine learning models trained on this data set.

\subsection{Manual annotation}
\label{sec:manual_annotation}
We developed a tool to perform region and dot annotation of panicles. Dot annotation consists of the user just clicking each panicle once. For region annotation, we start with a Simple Linear Iterative Clustering (SLIC) superpixel segmentation \cite{Achanta:10, Achanta:12}, and a few clicks per panicle (in our application , about $3$) are needed for annotation. Our annotation tool provides for $3$ sizes of superpixels to allow the user to choose how accurately to annotate the panicles.  Appendix \ref{sec:superpix_seg} has an example image with the three corresponding superpixel segmentation sets. Our tool also incorporates a feedback mechanism, where it will `guess' the panicle superpixels based on the current detection model, and the user only has to correct its predictions. This feedback loop becomes quite helpful for accelerating human annotation as well as model training. 

\begin{figure}[h]
  \begin{center}
  \includegraphics[width=\linewidth]{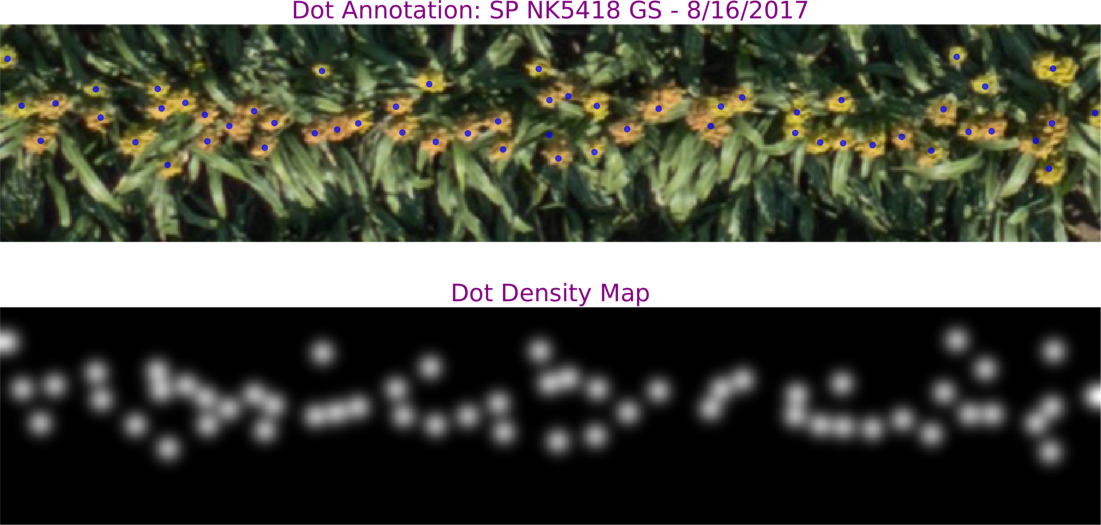}
  \end{center}
  \caption{Example dot annotation and corresponding dot density map.}\label{fig:dot}
\end{figure}

\begin{figure}[h]
  \begin{center}
  \includegraphics[width=\linewidth]{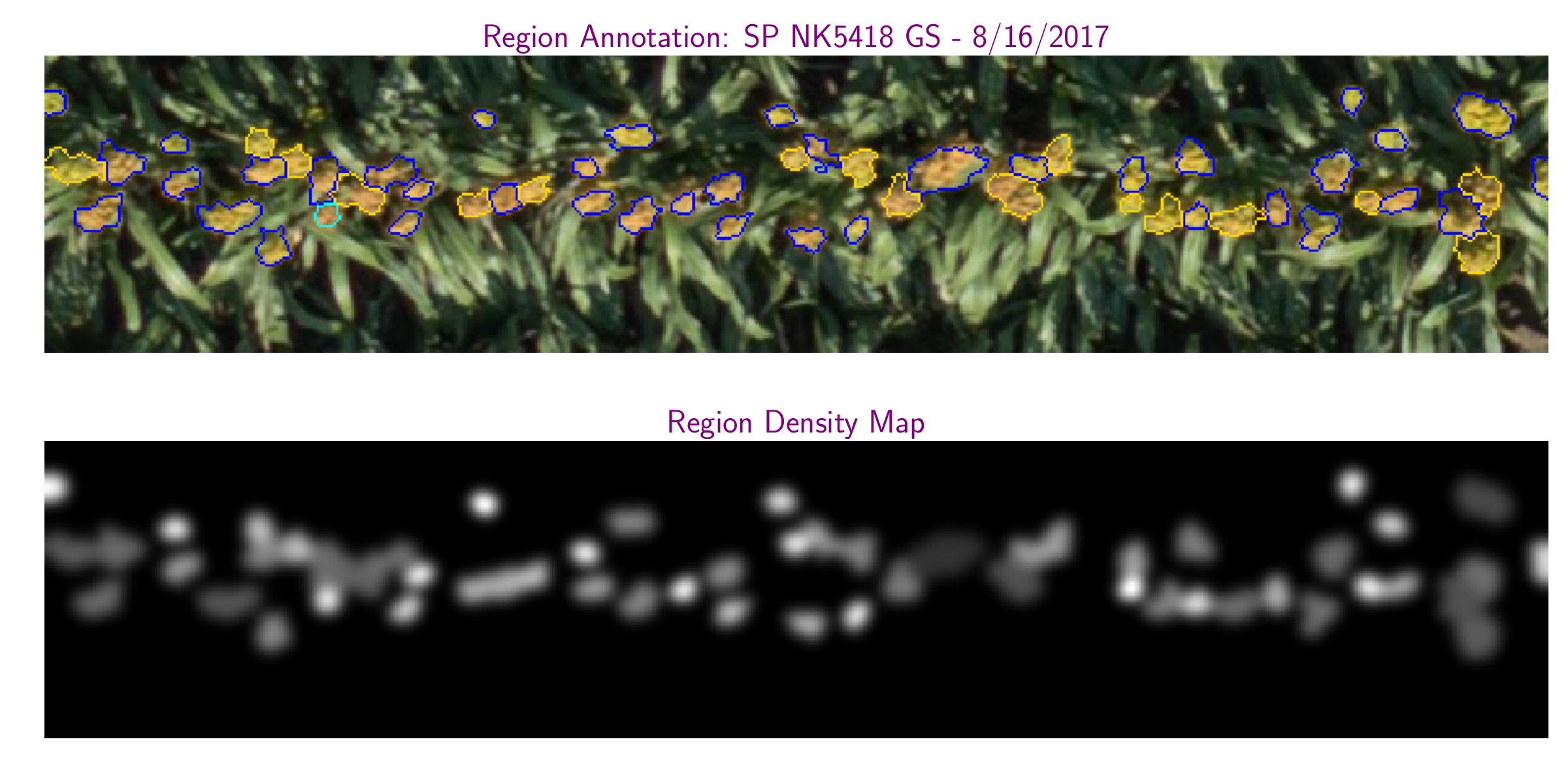}
  \end{center}
  \caption{Example region annotation and corresponding region density map.}\label{fig:area}
\end{figure}

The total number of panicles in all the images is 12,099. The annotated panicles corresponded to 35,329 superpixels for region annotation. We give some examples of easy and hard to annotate images in Appendix \ref{sec:example_images}.

% \subsection{Image Head Counts}
%  Nonetheless, the problem is difficult due to self occlusion later in the season.  Occlusion is particularly severe in grassy varieties.  Hence, to develop a good panicle detector and counting system requires a diverse, accurately annotated data set, and robust machine learning models trained on this data set.

% \karthi{This section tends to say that we have a simple data set, and a simple problem - we need to bring out the challenges better. One of the key things in our method seems to be our ``simultaneous detection and counting'' method, which seems quite novel. Usually people detect and then count, or just count without a way to do good detection --- we can do both simultaneously. For us good detection is crucial because identifying panicle boundaries has value in plant breeding.}

\subsection{Image Density Map Estimation}
\label{sec:density_maps}
The target for our regression network is density maps derived from annotations of panicle images. From the dot and region annotations, we respectively derive the dot and the region density maps. In the dot density map, we start with an indicator map that denotes the location of each panicle with a dot, and convolve a fixed width Gaussian kernel with it. This can be seen in Figure~\ref{fig:dot}. In the region density map, the user annotates the entire region of the panicle. The area under each distinct panicle region sums to 1. This is convolved with a small fixed width Gaussian (see Figure~\ref{fig:area}). For each of these density maps the total density equals the number of panicles in the image. The counting performance in using dot vs. region annotation was almost the same. However, the predicted dot density maps are also ideal for locating the center of the panicle, while the predicted region density map indicates the full extent of a panicle. The region density maps are also used along with the proposed instance segmentation approach, and this is lot harder to achieve with dot density maps.

% \karthi{In the results, are we comparing with logistic regression + color histograms?}

\section{Proposed Counting Regressor}
\label{sec:cnn_arch}
Due to one third of the images containing none or very few panicles, the CCNN, MCNN and CSRNet models (see Appendix \ref{sec:cnn_arch_addl} for the architecture diagrams) do not train properly on our data set.  In most training runs the model eventually converges to zero due to the nature of the data.  Adding a batch normalization layer after every convolutional layer fixed this issue and allowed all the models to be trained on our data set.  Conversely, adding batch normalization to the crowd counting data did not improve the performance.  This meant initializing CSRNet with VGG16\_bn -- a version of VGG16 with batch normalization.

Our proposed counting regressor was a tuned version of the CCNN approach \cite{Onoro:16} that has different number of layers and convolutional filter sizes.  To enable training from small data sets, with only hundreds of images, it is necessary to use an intermediate lifted high dimensional target. We use the image density map that distributes the count as a unit density for each panicle. Since our images only has small perspective distortions, we found that we do not need more complex architectures such as MCNN. In fact, the more complex architectures tend to hurt the performance, as we added enhancements to the CNN models.

The architecture we employed is shown in Figure~\ref{fig:system}.  We experimented with various choices of hyper parameters of the network, such as the number of layers, the size of the receptive filters or convolutional kernels (\texttt{kernel}), and the number of output channels (\texttt{dim}).  For the canonical CCNN we had \texttt{dim}=[32, 32, 32, 1000, 400, 1], \texttt{kernel}=[7, 7, 3, 1, 1, 1], whereas our best model used \texttt{dim}=[20, 40, 100, 400, 800, 1500, 100, 1], \texttt{kernel}=[13, 9, 5, 5, 5, 3, 1, 1].
 It was beneficial to make the number of filters be 1 for the last layer(s) so as to get a fully connected convolutional layer. The network had two max pooling layers inserted before the second and third convolutional layers.  As the max-pooling reduces the size of the output, the resulting image density map had to be down-sampled accordingly.  This problem can be potentially overcome using fully convolutional regression networks \cite{xie2018microscopy}. 

% \textit{This is one drawback with the architecture in our setting. We would like to train deep architectures, but then more max pooling layers are needed which leads to loss in density map resolution.}

We additionally enhanced our system, CCNN, MCNN and CSRNet through four separate mechanisms. First, we encode rotation and reflection invariance through median averaging of the CNN predictions. Secondly, we use thermal time \cite{Ritchie:91} as a proxy to encode the plant developmental stages. Thirdly, the time series of panicle count predictions is forced to monotonically increase through the use of isotonic regression \cite{DeLeeuw:77}.
Finally we used a panicle pixel detection map as an added input channel to the CNNs.  This detection map was generated using the system in Figure~\ref{fig:system} and as this system is quite accurate no noticeable difference was seen when using another CNN architecture to produce the detection map.

These enhancements are described in the rest of this section, and the corresponding results are included in Table~\ref{table:cnn}.

\subsection{Rotational Invariance}
Similar to training data augmentation, we also rotate and flip the data at test time and then use a statistic of all the predictions to create an invariant prediction. Both the mean and median are candidates for the invariant statistics, but we found the median to be the superior choice as it consistently yielded slightly better results due to its robustness to outliers. 
%Figure~\ref{fig:rotations} shows a concrete example of an image, the group of rotations by 90 degrees and flips and the corresponding predictions.  
%Table~\ref{table:rotations} gives the prediction errors for the dot and region annotation both for the mean and median statistics for the same system as in Table~\ref{table:cnn}.  

% \begin{table}[h]
% \newcolumntype{x}{>{\columncolor{limegreen!30}}l}
% \newcolumntype{s}{>{\columncolor{limegreen!60}}l}
% \newcolumntype{y}{>{\columncolor{melon!30}}c}
% \newcolumntype{t}{>{\columncolor{melon!60}}c}
% \newcolumntype{z}{>{\columncolor{blue!20}}c}
% \newcolumntype{u}{>{\columncolor{blue!40}}c}
% \begin{center}
% \begin{tabular}{xx||yy|}
%   \multicolumn{1}{s}{\bf Statistic} & \multicolumn{1}{s||}{Density Map} &
%   \multicolumn{1}{t}{MAE} &
%   \multicolumn{1}{t|}{$R^2$} \\ 
%   mean & dot & 1.56 &  0.968 \\
%   median &  & {\bf 1.55} & 0.968\\ 
%   mean & region & 1.48 & 0.974 \\
%   median & & {\bf 1.47} & 0.974 \\
% \end{tabular}

% \end{center}
%   \caption{CNN prediction errors for the mean and median statistics for the 8 rotated predictions.  The CNN has \texttt{dim}=[3, 20, 40, 20, 10], \texttt{kernel}=[9, 5, 5, 5].}\label{table:rotations}
% \end{table}

\subsection{Isotonic Regression}
\begin{figure*}[t]
  \begin{center}
  \includegraphics[width=\linewidth]{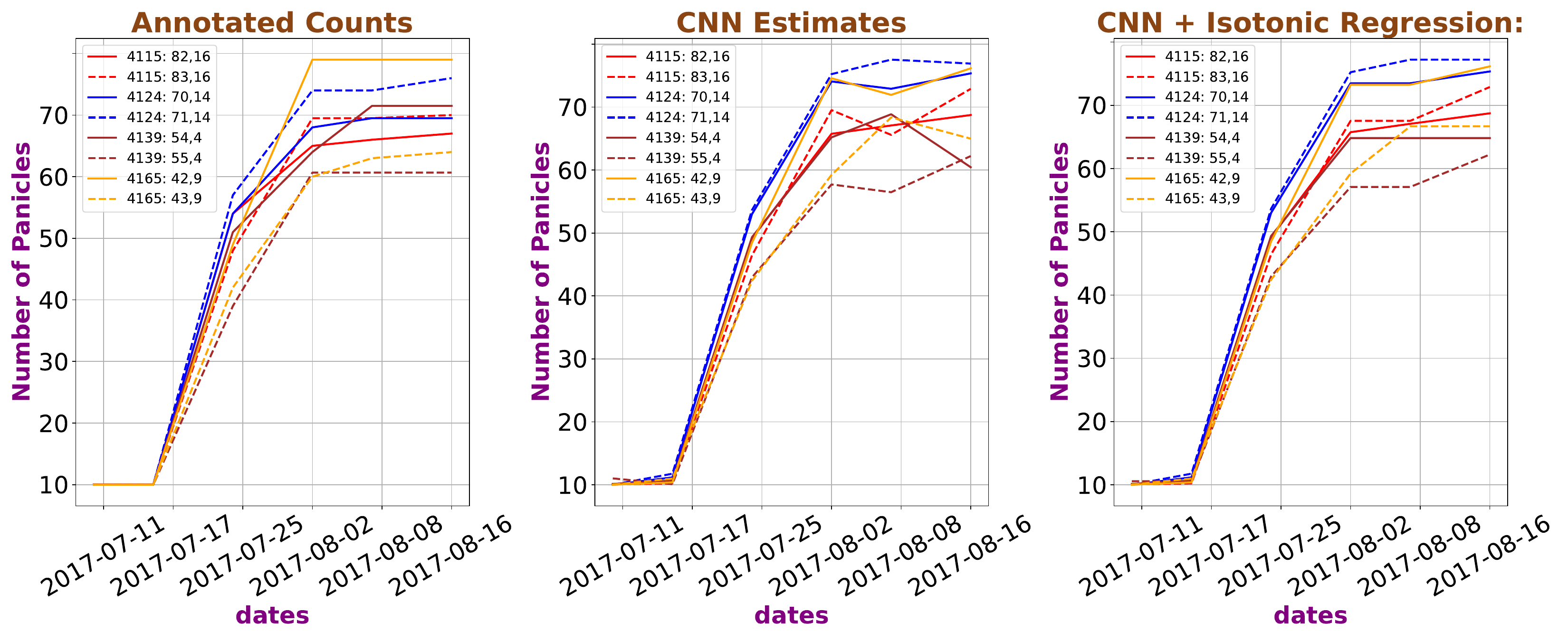}
  \end{center}
  \caption{Panicle predictions for all annotated row segments of the variety \texttt{SP NK5418 GS}.  Rows on different plots are in different colors.  The numbers in the legend are respectively the plot and the $x,y$ identifier of the row-segment.  The left plot shows counts from the human annotation, the middle figure shows predictions with the region density map CNN with median averaging.  The right plot shows the isotonic regression derived from the middle plot.  }\label{fig:isotonic}
\end{figure*}

The number of panicles should be a monotonically increasing function with time short of a drastic event (water lodging, diseases, extreme weather).  In fact this growth curve is roughly sigmoidal \cite{bartel1938growth}. Modelling errors on the other hand frequently lead to non-monotonic behavior as can be seen in the middle plot in Figure~\ref{fig:isotonic}. Correcting these anomalies should lead to improved performance.

Isotonic regression \cite{Chakravarti:89,DeLeeuw:77} forces the count to be non-decreasing through the least squares minimization
\begin{equation}\label{eq:isotonic}
\min_{c_i} \sum_{i=1}^n \|c_i - \hat{c}_i \|_2^2 \text{ subject to } c_1\leq
c_2\leq \cdots \leq c_n, 
\end{equation}
where $\hat{c}_i$ is the count estimates and $c_i$ is the desired monotonic correction. We use the efficient pool adjacent-violators algorithm (PAVA) \cite{Mair:09}, implemented in Python's \texttt{scikit-learn} library to solve (\ref{eq:isotonic}). Figure~\ref{fig:isotonic} shows the human annotation derived counts, the CNN predictions and the isotonic corrections side by side for one of the 18 varieties. In Appendix \ref{sec:best_res}, we show these results for our best system on all varieties.

% \begin{table}[h]
% \newcolumntype{x}{>{\columncolor{limegreen!30}}l}
% \newcolumntype{s}{>{\columncolor{limegreen!60}}l}
% \newcolumntype{y}{>{\columncolor{melon!30}}c}
% \newcolumntype{t}{>{\columncolor{melon!60}}c}
% \newcolumntype{z}{>{\columncolor{blue!20}}c}
% \newcolumntype{u}{>{\columncolor{blue!40}}c}
% \begin{center}
% \begin{tabular}{xx||yy|}
%   \multicolumn{1}{s}{\bf Statistic} & \multicolumn{1}{s||}{Density Map} &
%   \multicolumn{1}{t}{MAE} &
%   \multicolumn{1}{t|}{$R^2$} \\ 
%   mean & dot & 1.48 &  0.973 \\
%   median &  & 1.47 & 0.973\\ 
%   mean & region & 1.40 & 0.977 \\
%   median & & {\bf 1.40} & 0.977 \\
% \end{tabular}
% \end{center}
%   \caption{CNN prediction errors for isotonic regression.  The CNN has \texttt{dim}=[3, 20, 40, 20, 10], \texttt{kernel}=[9, 5, 5, 5].}\label{table:isotonic}
% \end{table}

\subsection{Thermal Time}
\label{sec:thermal_time}
As the plant develops it eventually reaches a stage where panicles appear for some time until they are fully developed.  The temperature is a very important factor in plant development \cite{Ritchie:91} - a plant is stressed when it is cold and grows faster when it is warm.  The plant development correlates well with the thermal time, measured as growing degree days (GDD).  Specifically for Sorghum, we define the GDD as the number of degree days above $50^{\circ}\mathrm{F}$ since planting.  With this definition a short season hybrid for example needs on average $1848^{\circ}\mathrm{F}$ to reach flowering.  Using the thermal time in place of the time of the year also allows us to use the same model in a different location or for a different year.  There are many ways to provide the thermal time to the CNN, but we provided it as a separate channel for the image as in Figure~\ref{fig:system}. 

% \begin{figure}[h]
%   \begin{center}
%   \includegraphics[width=\linewidth]{figures/thermal.pdf}
%   \end{center}
%   \caption{A thermal time based CNN.}
%   \label{fig:thermal}
% \end{figure}

\subsection{Detection Map}
\label{sec:detection_map}
We used our CCNN like system with \texttt{dim}=[20, 40, 100, 400, 800, 1500, 100, 1], \texttt{kernel}=[13, 9, 5, 5, 5, 3, 1, 1] with the binary panicle pixel mask as a regressor.  We refer to the predicted mask with values ranging from 0-1 as the detection map and added it to the input data channels.  The detection map as well as the thermal time allows the CNN models to ignore the parts of the images that do not contain panicles.  This results in the models effectively being trained on a much smaller data set, which meant that more complex models saw less benefit (if any) from these added channels.

\subsection{Experimental Results}
Table~\ref{table:cnn} shows the results for the CCNN, MCNN, CSRNet and our own tuned version of the CCNN architecture.  It can be seen that the models with the largest number of parameters are MCNN and CSRNet and they do not see nearly the same gain as the smaller models when adding the thermal and panicle detection channels.  CSRNet in particular sees no gain despite being smaller than MCNN.  This is due to CSRNet being a much deeper architecture that suffers from a larger degree of over training (as discussed in \cite{li2018csrnet}).  We observed that the mean absolute error for the CSRNet model is 0.29 for training data vs. 1.19 on test data for the region density model.  The new channels allows CSRNet to effectively focus on only the panicle data (less than 2\% of the image data), so the over training data is exacerbated.  Our model with all the bells and whistles gives an MAE of 1.06 compared to 1.17 for the plain CSRNet model - a relative improvement of nearly 10\%.  We expect that as the amount of annotated data increases the larger models will also benefit from the thermal and detection map channels.

\begin{table*}[htbp]

\newcolumntype{x}{>{\columncolor{limegreen!25}}l}
\newcolumntype{f}{>{\columncolor{limegreen!45}}l}
\newcolumntype{s}{>{\columncolor{limegreen!60}}l}
\newcolumntype{y}{>{\columncolor{melon!25}}c}
\newcolumntype{g}{>{\columncolor{melon!45}}c}
\newcolumntype{t}{>{\columncolor{melon!60}}c}
\newcolumntype{z}{>{\columncolor{blue!20}}c}
\newcolumntype{h}{>{\columncolor{blue!30}}c}
\newcolumntype{u}{>{\columncolor{blue!45}}c}

  \begin{center}
  \small
    \begin{tabular}{xx||yyyyy||zzzzz}
    \multicolumn{1}{s}{System} & \multicolumn{1}{s||}{Size} & 
    \multicolumn{1}{t}{Base} & \multicolumn{1}{t}{+Rot} & \multicolumn{1}{t}{+Isot} &
    \multicolumn{1}{t}{+ther} & \multicolumn{1}{t||}{+det} &
    \multicolumn{1}{u}{Base} & \multicolumn{1}{u}{+Rot} & \multicolumn{1}{u}{+Isot} &
    \multicolumn{1}{u}{+ther} & \multicolumn{1}{u}{+det} \\
    \multicolumn{2}{s||}{} & \multicolumn{5}{t||}{region} & \multicolumn{5}{u}{dot}\\
\hline
    CCNN & 2MB & 1.52 & 1.46 & 1.38 & 1.26 & 1.14 & 1.62 & 1.53 & 1.39 & 1.33 & 1.11 \\
    Ours & 21MB & 1.38 & 1.36 & 1.28 & 1.17 & {\bf 1.08} & 1.39 & 1.38 & 1.28 & 1.18 & {\bf\color{red} 1.06} \\
    MCNN & 86MB & 1.43 & 1.37 & 1.29 & 1.29 & 1.21 & 1.38 & 1.38 & 1.28 & 1.25 & 1.24 \\
    CSRNet & 62MB & 1.19 & 1.14 & 1.11 & 1.09 & 1.10 & 1.17 & 1.12 & 1.09 & 1.10 & 1.11 \\
    \end{tabular}%
\end{center}

  \caption{CNN prediction errors for counting with various enhancements. We use the abbreviations "Rot" for rotational invariance, "Isot" for isotonic regression, "ther" for thermal time and "det" for detection map.}
  \label{table:cnn}%
\end{table*}%

\subsection{Results on Publicly Available Data}
The only publicly available data we are aware of for counting panicles in Sorghum aerial imagery was released with \cite{guo2018aerial} by Guo \textit{et al.} In this section, we provide a brief description of the dataset, and also the results for counting using our proposed regressor.

Guo \textit{et al.}'s data contains a training set of 40 large images with roughly 105 panicles in each image and two test sets. The test sets are referred to as dataset 1 and dataset 2. Dataset 1 has 489 panicles per image on average, while dataset 2 typically has 106 panicles per image on average. The resolution is higher than for our images (0.45cm $\times$ 0.45cm per pixel versus 0.66cm $\times$ 0.66cm per pixel).

The data is all from one date, so we cannot take advantage of the thermal time or isotonic regression that gave us some gain in performance. Also, experiments using a detection layer as an input channel did not yield gain in performance. We do not know exactly why this is, but suspect that both the region density map as well as the detection map are less effective due to the larger extent of each panicle resulting from the increase in resolution. Also, we do not know the accuracy of the pixel detection density map as the test sets were only annotated in terms of panicle centers. Table~\ref{table:GuoMAE} gives the results for our proposed counting regressor model, CCNN, MCNN and CSRNet. CSRNet is superior to the other methods and perhaps the reason is that the dilated convolutions gives the model a bigger effective receptive field -- hence the ability to grapple with the larger extent of the panicles.
In  \cite{guo2018aerial} the authors do not report MAE, but rather report the coefficient of determination $R^2$.  We give $R^2$ results to compare with their results in Table~\ref{table:GuoR2}.

\begin{table}[htbp]

\newcolumntype{x}{>{\columncolor{limegreen!25}}l}
\newcolumntype{f}{>{\columncolor{limegreen!45}}l}
\newcolumntype{s}{>{\columncolor{limegreen!60}}l}
\newcolumntype{y}{>{\columncolor{melon!25}}c}
\newcolumntype{g}{>{\columncolor{melon!45}}c}
\newcolumntype{t}{>{\columncolor{melon!60}}c}
\newcolumntype{z}{>{\columncolor{blue!20}}c}
\newcolumntype{h}{>{\columncolor{blue!30}}c}
\newcolumntype{u}{>{\columncolor{blue!45}}c}

  \begin{center}
  \small
    \begin{tabular}{xx||yyyy}
    \multicolumn{1}{s}{System} & \multicolumn{1}{s||}{Density} & \multicolumn{2}{t}{Dataset 1} & \multicolumn{2}{t}{Dataset 2}\\
     \multicolumn{1}{s}{} & \multicolumn{1}{s||}{} & \multicolumn{1}{t}{Base} & \multicolumn{1}{t}{+rot} & 
   \multicolumn{1}{t}{Base} & \multicolumn{1}{t}{+rot} \\
\hline
    CCNN & dot    & 21.67 & 22.94 & 5.00 & 4.24 \\
         & region & 19.56 & \bf 19.37 & 4.16 & 4.10 \\
    Ours & dot    & 20.39 & 19.52 & 3.23 & 3.15 \\
         & region & 23.64 & 23.46 & \bf 3.56 & 3.66 \\
    MCNN & dot & 20.61 & 20.14 & 3.72 & 3.29 \\
         & area & 22.52 & 23.85 & 3.16 & 3.32 \\
    CSRNet & dot    & 17.12 & {\color{red}\bf 16.70} & 2.72 & {\color{red}\bf 2.20} \\
           & region & 21.32 & 20.82 & 3.84 & 3.57 \\
    \end{tabular}%
\end{center}
  \caption{MAE on the two test sets released with \cite{guo2018aerial} for four different CNN architectures for counting.}
  \label{table:GuoMAE}%
\end{table}%
\begin{table}[htbp]

\newcolumntype{x}{>{\columncolor{limegreen!25}}l}
\newcolumntype{f}{>{\columncolor{limegreen!45}}l}
\newcolumntype{s}{>{\columncolor{limegreen!60}}l}
\newcolumntype{y}{>{\columncolor{melon!25}}c}
\newcolumntype{g}{>{\columncolor{melon!45}}c}
\newcolumntype{t}{>{\columncolor{melon!60}}c}
\newcolumntype{z}{>{\columncolor{blue!20}}c}
\newcolumntype{h}{>{\columncolor{blue!30}}c}
\newcolumntype{u}{>{\columncolor{blue!45}}c}

  \begin{center}
  \small
    \begin{tabular}{xx||yyyy}
    \multicolumn{1}{s}{System} & \multicolumn{1}{s||}{Density} & \multicolumn{2}{t}{Dataset 1} & \multicolumn{2}{t}{Dataset 2}\\
     \multicolumn{1}{s}{} & \multicolumn{1}{s||}{} & \multicolumn{1}{t}{Base} & \multicolumn{1}{t}{+rot} & 
   \multicolumn{1}{t}{Base} & \multicolumn{1}{t}{+rot} \\
\hline
    Quadratic SVM \cite{guo2018aerial} & - & 0.84 & & 0.56 & \\
    % RetinaNet & - & 0.82 & & 0.77 & \\
    CCNN & dot    & 0.88 & 0.86 & 0.52 & 0.61 \\
         & region & 0.86 & \bf 0.86 & 0.63 & 0.65 \\
    Ours & dot    & 0.90 & {\color{red}\bf 0.90} & 0.77 & 0.80 \\
         & region & 0.85 & 0.84 & 0.80 & \bf 0.82 \\
    MCNN & dot & 0.81 & 0.80 & 0.73 & 0.79 \\
         & region & 0.64 & 0.57 & 0.79 & 0.76 \\
    CSRNet & dot    & 0.89 & {\color{red}\bf 0.90} & 0.86 & {\color{red}\bf 0.88} \\
           & region & 0.84 & 0.85 & 0.73 & 0.75 \\
    \end{tabular}%
\end{center}
  \caption{Coefficient of determination, $R^2$, on the two test sets released with \cite{guo2018aerial} for four different CNN architectures for counting as well as the Quadratic SVM method proposed in \cite{guo2018aerial}.}
  \label{table:GuoR2}%
\end{table}%

\section{Panicle Detection and Segmentation}
\label{sec:detect_segment}
% The amount of data collected for panicle detection is relatively modest compared to standard computer vision data sets.  Sorghum is also just one of many crops and one can imagine a plethora of computer vision data sets being needed for the new era of UAV assisted agronomy.  In order to facilitate rapid collection of annotated data we experimented with using our CNN model to do a first pass annotation that can then be corrected by the human annotator.  The automatic annotation tool can operate on two levels according to the accuracy of the model.  First we can simply do a panicle detector that operates as a foreground-background segmentation.  The annotator would then have to separate panicles wherever there is occlusion.  The second level is to automatically separate the panicles wherever there is occlusion as well.  Then the annotator would only have to correct errors.

% Since the annotation tool operates in terms of superpixels we target superpixel level segmentations and not pixel level segmentations.  Let's first look at the panicle detector.

The size and shape of a panicle is by itself an interesting phenotype, but to derive this information we need to move beyond the count and estimate a panicle instance segmentation indicating which pixels belongs to each panicle versus the background. We propose a novel \textit{instance segmentation} approach wherein we: (a) use the image detection map to directly detect panicle superpixels, and (b) group superpixels into panicle segments using clustering and the region density map.  
Early in the growth season, when there is no occlusion, segmentation can be done through the connected components from the panicle detection map.  The late season images can have severe occlusion problems - especially for grassy varieties - and the difficulty is increased by overlapping panicles with homogeneous texture and few discernible edges.  
 We use the fact that the integrated density over a particular panicle should equal $1$ and rely on the region density map to extract individual panicles.  We then perform density aware greedy clustering to obtain panicle segments.

% this paragraph can be cut if we need space
%While we do not claim that the accuracy of this instance segmentation is state-of-the-art, we find it very interesting that an instance segmentation can be derived directly from the region density map.  The classical methodology to count objects in images is to first detect/segment and then to count.  Here we reverse the order by first counting (through regression) and then forming the instance segmentation via density aware clustering.  The performance of our instance segmentation system is naturally bounded by the performance of the underlying CNN used to count panicles.

\subsection{Detecting Panicle Superpixels}
In order to have efficient algorithms we cluster superpixels in place of pixels.  We used the CNN in Figure~\ref{fig:system} with the panicle detection map as a regressor.  At test time this CCNN model will predict a detection map, thus acting as a panicle pixel detector (see Appendix \ref{sec:seg_det_cnn} for more details).  The predicted detection map can be interpreted as an estimated pixel panicle probability.  To convert this into a superpixel panicle detector we used the mean panicle pixel probability as the superpixel panicle probability.  We set a probability threshold $\alpha$ as a detection threshold and varied $\alpha$ between 0.3 and 0.6 to affect the precision-recall balance. The panicle superpixel detector algorithm is given in Algorithm \ref{algo:superpix_det}. Note that $S$ is the superpixel map (matrix) that assigns a pixel to a superpixel value.

\begin{algorithm}
\DontPrintSemicolon
  \KwInput{thermal time \texttt{tt}, image $I$}
  \KwOutput{Set of panicle superpixels $P$}
  $D=\mathrm{detection\_CNN}(\texttt{tt},I)$
  
  \tcc{$D(i,j)=$ panicle pixel probability.}
  $S = \mathrm{SLIC}(I)$ \tcp*{$S(i,j)=t$ for superpixel $t$.}
  $$ P = \left\{ p : \frac{\sum_{ij:S(i,j)=p}D(i,j)}{\sum_{ij:S(i,j)=p} 1} \geq \alpha \right\} $$
\caption{Panicle Superpixel Detector}
\label{algo:superpix_det}
\end{algorithm}

\subsection{Cluster Fitness}
Discovering which superpixels belong to a single panicle is difficult in the presence of occlusion as they have the texture and color tends to be the same.  This leads us to focus the objective on the shape-compactness and to depend to a large degree on the region density map.  This lead us to use the definition of cluster fitness defined in Algorithm~\ref{algo:fitness}.  Many clustering algorithms are distance based (K-means, DBSCAN, Agglomerative Clustering), but can also be formulated in terms of cluster fitness.  Our cluster fitness is a weighted combination of how small the cluster variance is and how close the total cluster density mass is to $\beta$.  $\beta$ should ideally be 1, but we change it to vary the number of recovered panicles for the precision-recall curve.  $C$ in the algorithm is a map from superpixel panicle indices to clusters.

\begin{algorithm}
\DontPrintSemicolon

  \KwInput{cluster $c$, $\beta$, superpixel map $S$, \texttt{tt}, cluster assignment $C$}
  \KwOutput{cluster fitness $f$}
  $R=\mathrm{region\_CNN}(\texttt{tt},I)$ \tcp*{Region density map}
  \For{$i,j$}{
  \tcc{$v$ - vectors for clustering pixels}
  $v(i,j):=[\gamma i,\gamma j,\mathrm{red}, \mathrm{green}, \mathrm{blue}]$\tcp*{$\gamma=10$}
  
  }
  $n = \sum_{i,j:C(S(i,j))=c} 1$\tcp*{pixel count}
  mean = $\frac{1}{n}\sum_{i,j:C(S(i,j))=c} v(i,j)$\\
  var = $\frac{1}{n}\sum_{i,j:C(S(i,j))=c} (v(i,j)-\mathrm{mean}(c))^2$\\
  $d = \sum_{i,j:C(S(i,j))=c} R(i,j)$\tcp*{density mass}
  $f =  n(\delta (d-\beta)^2 + \mathrm{var})$ \tcp*{$\delta=46775$, $\beta\approx 1$. }
  \KwRet{$f$}
\caption{Cluster Fitness}\label{algo:fitness}
\end{algorithm}

\subsection{Instance Segmentation}
\begin{figure}[h]
  \begin{center}
  \includegraphics[width=.9\linewidth]{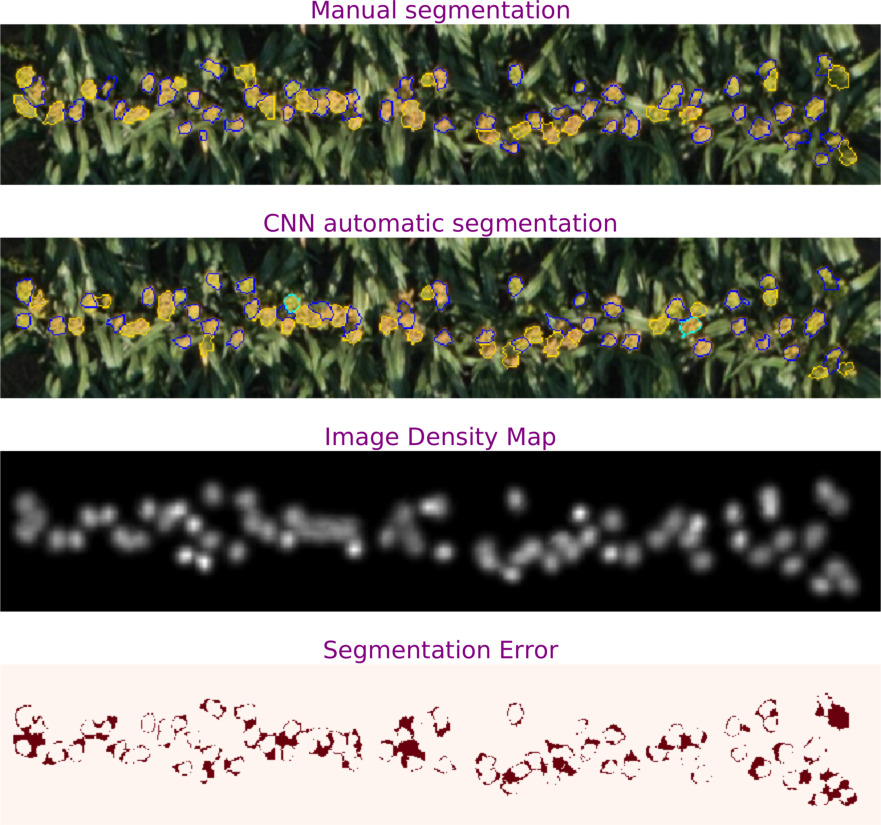}
  \end{center}
  \caption{A comparison of the manual and automatic segmentation.}\label{fig:segmentation}
\end{figure}

To discover the individual panicle segments we used greedy bottom-up clustering.  Initially every panicle superpixel is its own cluster.  We then progressively merge pairs of clusters until the accumulated cluster fitness is increased.  At each step the pair with the best cluster fitness change is identified for the merge.   Algorithm~\ref{algo:cluster} shows the details.  A result of this clustering process can be seen in Figure~\ref{fig:segmentation}.
%which shows the manual segmentation, the automatic segmentation and the segmentation error.  The segmentation pin-pointed the panicles, but had a large false positive rate for the panicle detector.  This was needed to avoid creating holes in the segmentation.  

\begin{algorithm}
\DontPrintSemicolon

  \KwInput{$P$, $\beta$, $S$, \texttt{tt}}
  \KwOutput{clusters $C$}
  $C= P$ \tcp*{All superpixels are clusters}
  $\mathrm{loss}=-1$\\
  \While{$\mathrm{loss}<0$}{
  $B=(0,0,0)$\\ 
  \For(\tcp*[h]{find best merge}){$c_1\neq c_2\in\mathrm{set}(C)$}{
  %\For{$c_1\neq c_2\in\mathrm{set}(C)$}{
    \If{$\mathrm{connected}(c_1,c_2)$}{
        \tcc{cost of merging $c_1$ and $c_2$}
        $\mathrm{loss} = \mathrm{fitness}(c_1,c_2)-\mathrm{fitness}(c_1)-\mathrm{fitness}(c_2)$\\
        \If{$\mathrm{loss}<B(0)$}{
            $B=(\mathrm{loss},c_1,c_2)$
        }
    }
  }
  $\mathrm{loss},c_1, c_2 = B$\\
  \If{$\mathrm{loss}<0$}{
    \For(\tcp{merge $c_1,c_2$}){$p\in P$}{
        \If{$C(p)==c_2$}{
            $C(p) = c_1$
        }
    }
  }
  }
\KwRet{$C$}
\caption{Instance Segmentation}\label{algo:cluster}
\end{algorithm}

Computing the Intersection of Union (IoU) for instance segmentation not based on rectangle annotation is more involved.  To compute the IoU requires first an alignment procedure so as to not allow one panicle to correspond to two panicles in the annotation.  This problem is known as a transportation or assignment problem and can be solved with the Hungarian algorithm also known as the  Kuhn--Munkres  algorithm  or  Munkres  assignment  algorithm \cite{kuhn1955hungarian, kuhn1956variants, munkres1957algorithms}.  We varied $\alpha$ and $\beta$ (Algorithm~\ref{algo:cluster}) to compute a precision-recall curve from which we derive the mAP to be 0.66 for a detection threshold of IoU=0.5.  The performance of our instance segmentation procedure is bounded by the quality of the region segmentation system and it is possible to obtain better performance using a specialized approach. However, our methodology is extremely efficient and compatible with our counting algorithm. It can also be accessed real time in a GUI and requires little additional code once a counting CNN is provided.  

\section*{Acknowledgements}
This  work  was  supported  by  the  Advanced  Research  Projects  Agency Energy (ARPA-E), U.S. Department of Energy under Grant DE-AR0000593.
The authors acknowledge the contributions of the Purdue and the IBM project teams for field work, data collection, processing and discussions. They thank Prof. Mitchell Tuinstra and Prof. Clifford Weil for leading and coordinating the planning, experimental design, planting, management, and data collection portions of the project. The authors deeply appreciate the help and support provided by Prof. Melba Crawford, Prof. Edward Delp, Prof. Ayman Habib, Prof. David Ebert and their students Ali Masjedi, Dr. Zhou Zhang, Dr. Javier Ribera Prat, Yuhao Chen, Dr. Fangning He, and Jieqiong Zhao. They were instrumental in creating and providing the image-related data used in this paper and also provided important feedback at various stages. The authors thank Andrew Linvill for helping with manual annotations, and Prof. Addie Thompson for providing suggestions to improve the annotation tool and helping us interpret the ground truth data. Dr. Neal Carpenter and Dr. Naoki Abe are acknowledged for scientific discussions. Finally, the authors thank Dr. Wei Guo and Dr. Scott Chapman for releasing and providing the public Sorghum head dataset used in our comparisons, and for the scientific discussions.

{\small
\bibliographystyle{ieee}
\bibliography{dhc}
}
\clearpage
\newpage 

\appendix

\section{Segmentation Using a Panicle Detection CNN}
\label{sec:seg_det_cnn}
\begin{figure}[h]
  \begin{center}
  \includegraphics[width=\linewidth]{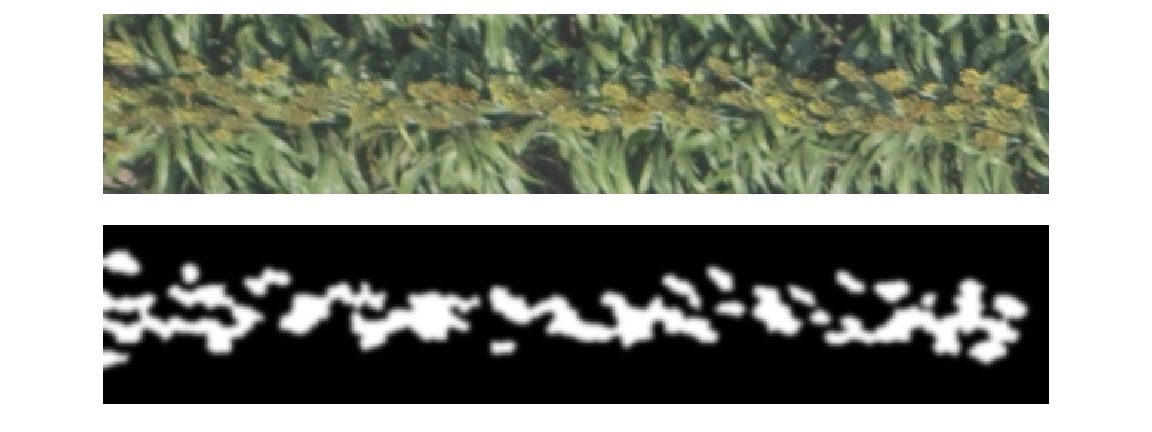}
  \end{center}
  \caption{An example regression target for the panicle detection CNN model.}\label{fig:fgbg}
\end{figure}

In order to paint a complete and fair picture we have to describe the CNN detection system.  We trained a CNN with the same architecture as the CCNN where the density map regression targets were replaced with a detection map.  Our detection map was a smooth version of the binary foreground-background indicator constructed by convolving against a fixed width gaussian the same way we did for the density map.  We see an example detection map in Figure~\ref{fig:fgbg}.

\section{Sorghum Pedigrees}
\label{sec:pedigrees}
The data collected consisted of images of the Hybrid Calibration Panel where the 18 hybrid varieties of sorghum listed in Table~\ref{table:pedigrees} were grown.

\begin{table}[h]
\newcolumntype{x}{>{\columncolor{limegreen!30}}l}
\newcolumntype{s}{>{\columncolor{limegreen!60}}l}
\newcolumntype{y}{>{\columncolor{melon!30}}l}
\newcolumntype{t}{>{\columncolor{melon!60}}c}
\newcolumntype{z}{>{\columncolor{blue!20}}c}
\newcolumntype{u}{>{\columncolor{blue!40}}c}
\begin{center}
\begin{tabular}{y}
  \multicolumn{1}{t}{\bf Pedigree} \\ 
  \texttt{PH 849F FS} \\
  \texttt{PH 877F FS} \\
  \texttt{RS 327x36 BMR FS} \\
  \texttt{RS 341x10 FG white} \\
  \texttt{RS 366x58 FG white} \\
  \texttt{RS 374x66 FS} \\
  \texttt{RS 392x105 BMR FS} \\
  \texttt{RS 400x38 BMR SG} \\
  \texttt{RS 400x82 BMR SG} \\
  \texttt{SP HIKANE II FS} \\
  \texttt{SP NK300 FS} \\
 \texttt{SP NK5418 GS} \\
 \texttt{SP NK8416 GS} \\
 \texttt{SP SS405 FS} \\
 \texttt{SP Sordan 79 FS} \\
 \texttt{SP Sordan Headless FS PS} \\
 \texttt{SP Trudan 8 FS} \\
 \texttt{SP Trudan Headless FS PS} 
\end{tabular}
\end{center}
  \caption{List of the 18 sorghum pedigrees grown in the 2017 hybrid calibration panel.
  }\label{table:pedigrees}
\end{table}

\section{Examples Images}
\label{sec:example_images}
\begin{figure}[h]
  \begin{center}
  \includegraphics[width=\linewidth]{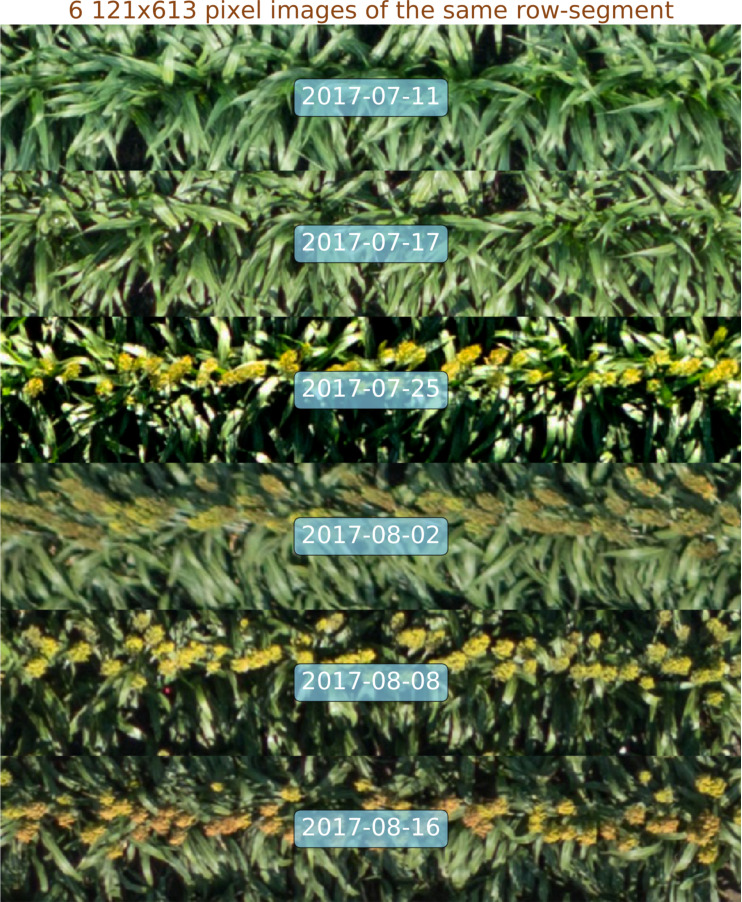}
  \end{center}
  \caption{Example Sorghum UAV images from the 2017 hybrid calibration panel.}\label{fig:data}
\end{figure}

Figure~\ref{fig:data} shows images over a row segment for the variety \texttt{SP NK5418 GS} for the six dates the UAV was flown.  Note that there is significant variations across the dates even though the variety is the same.

\begin{figure*}[h]
  \begin{center}
    \includegraphics[width=\linewidth]{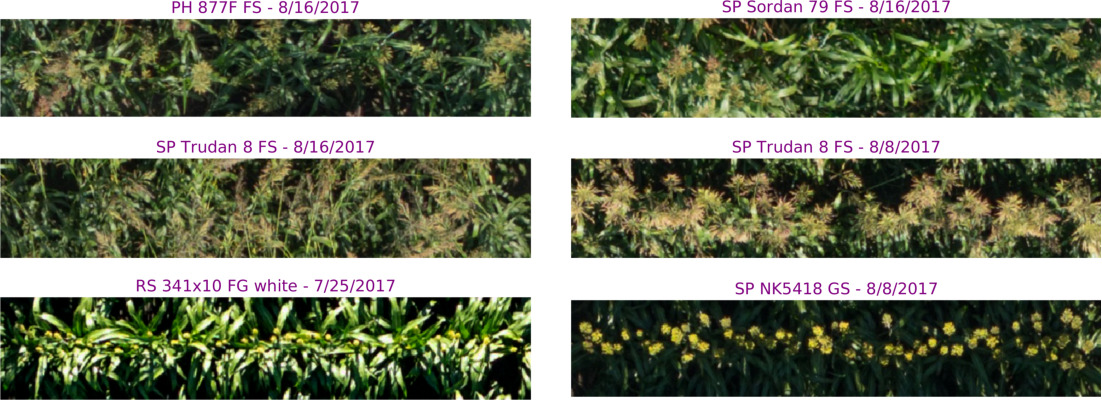}  
  \end{center}
  \caption{Example row-segment images where it is harder to annotate and count the panicles.}\label{fig:examples}
\end{figure*}

The 2017 hybrid calibration data contained a mix of images, some of which were easy and some of which were hard to annotate. Figure~\ref{fig:examples} show some example row-segment images in the data that exemplified the different problems encountered when trying to annotate the data.

\begin{figure}[h]
  \begin{center}
  \includegraphics[width=\linewidth]{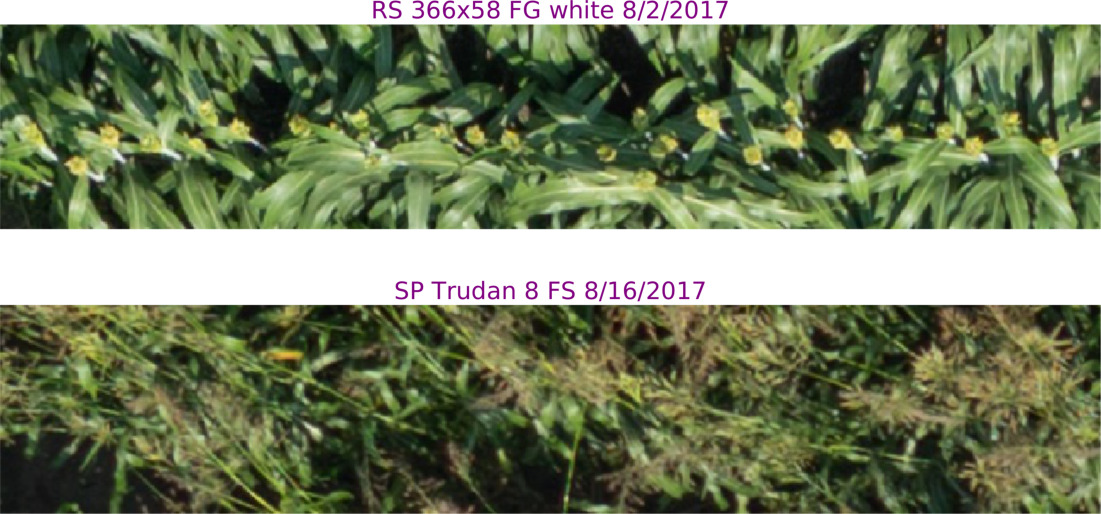}
  \end{center}
  \caption{Example of panicle size variation.}\label{fig:panicle}
\end{figure}

Figure \ref{fig:panicle} shows two varieties of sorghum where the panicle sizes are quite different.  The second image also demonstrates how dramatic the self-occlusion can be in the grass-like varieties of sorghum.

\begin{figure*}[t]
  \begin{center}
  \includegraphics[width=\linewidth]{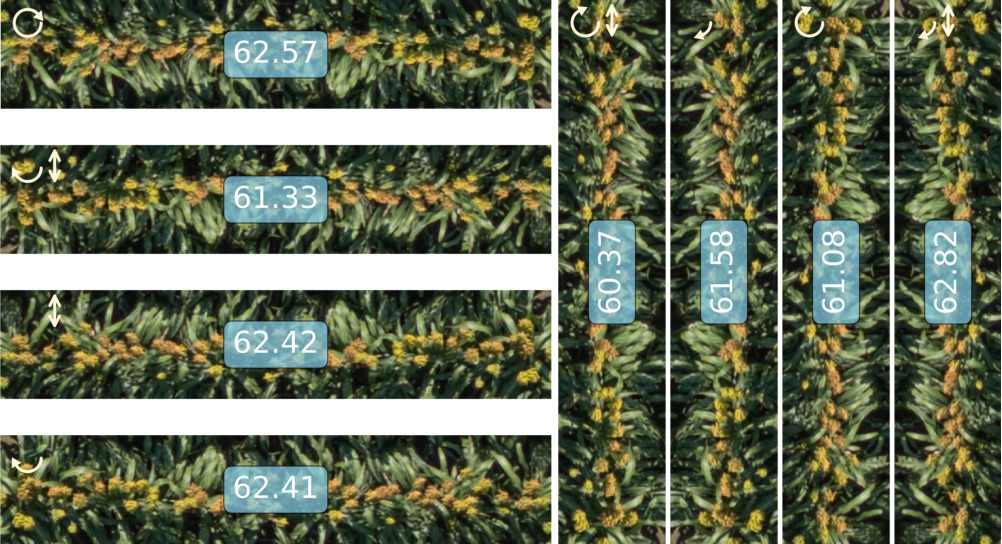}
  \end{center}
  \caption{An example of the CNNs predictions for flips and rotations for one image.  The manual human count was 57 for this image, while the median statistics gave 61.99 and the mean statistics gave 61.82.  The arrows indicate the rotation and the up down arrows a vertical flip.
  }\label{fig:rotations}
\end{figure*}

Figure~\ref{fig:rotations} shows a concrete example of an image, the group of rotations by 90 degrees and flips and the corresponding predictions. 

\begin{figure*}[t]
  \begin{center}
  \includegraphics[width=\linewidth]{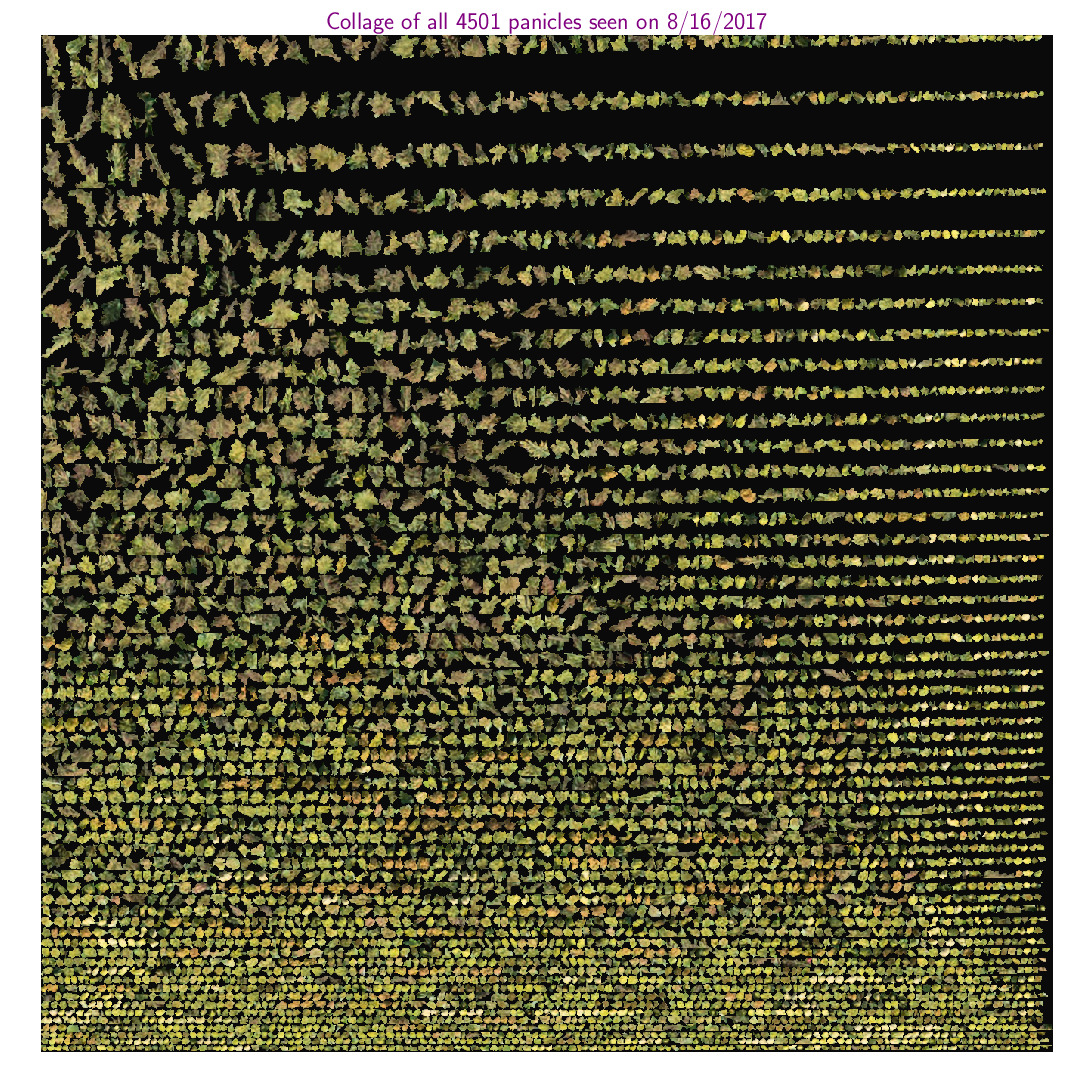}
  \end{center}
  \caption{A collage of all 4501 panicles seen on 8/16/2017.  
  }\label{fig:fullcollage}
\end{figure*}

Figure~\ref{fig:fullcollage} shows an image representation of all manually annotated panicles on 8/16/2017 on the hybrid calibration plot.  There were a total of 4501 panicles observed on the last date and these represent 17 out of the 18 varieties as heads of one variety only appeared in September.

\section{CNN Architectures}
\label{sec:cnn_arch_addl}
\begin{figure*}[t]
  \begin{center}
  \includegraphics[width=\linewidth]{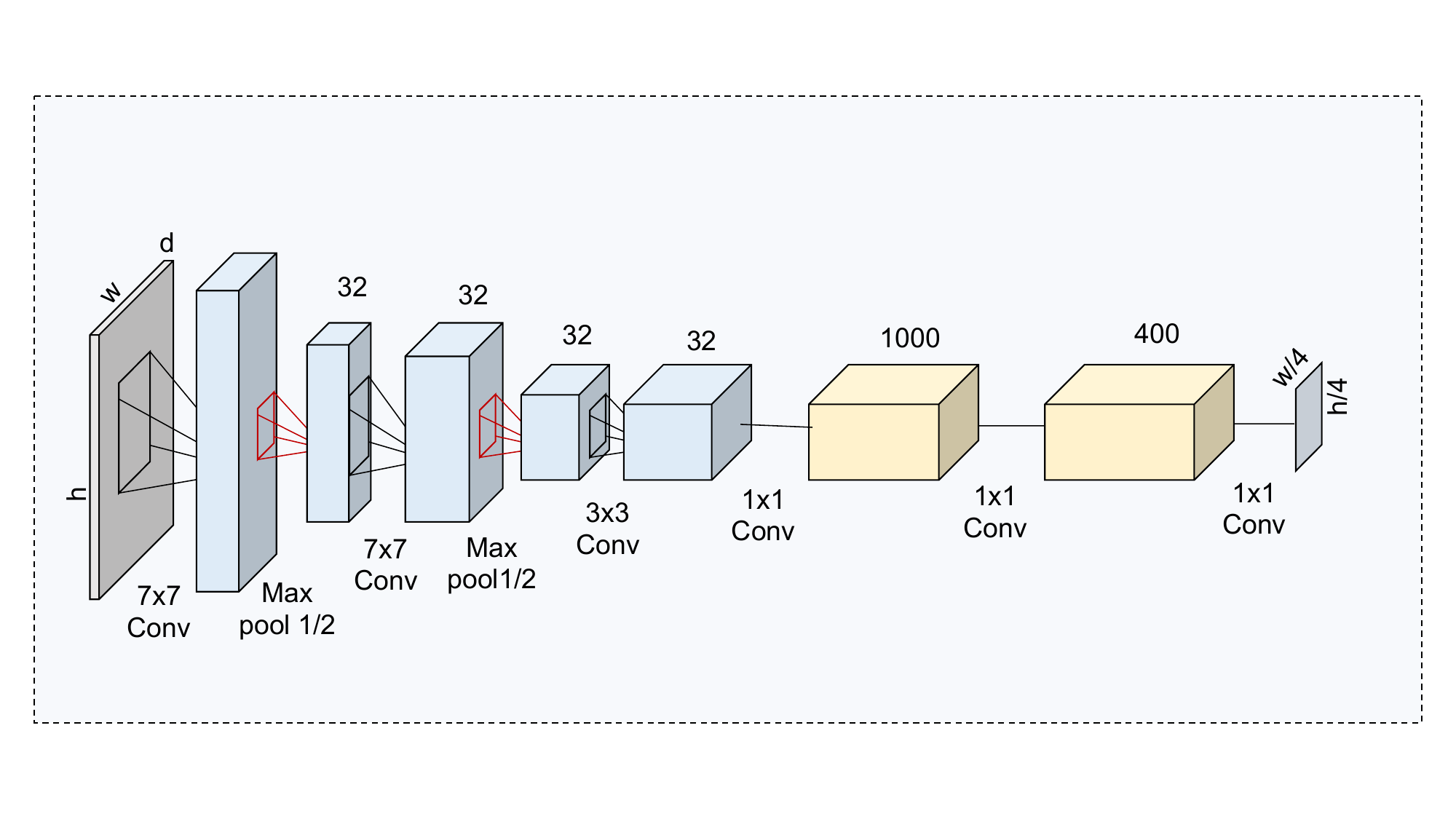}
  \end{center}
  \caption{The CCNN architecture. 
  }\label{fig:ccnn_arch}
\end{figure*}

\begin{figure*}[t]
  \begin{center}
  \includegraphics[width=\linewidth]{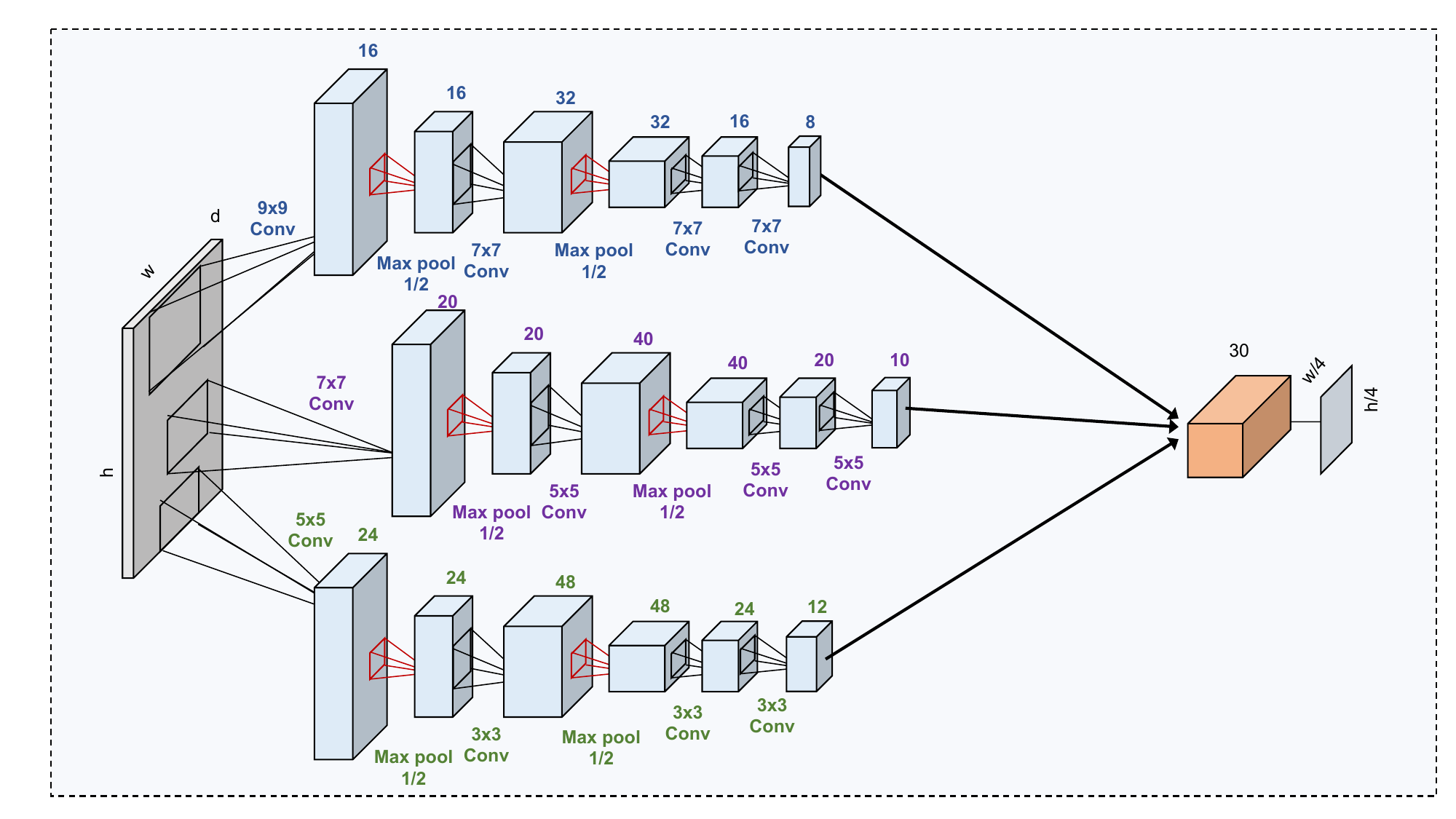}
  \end{center}
  \caption{The MCNN architecture  
  }\label{fig:mcnn_arch}
\end{figure*}

\begin{figure*}[t]
  \begin{center}
  \includegraphics[width=\linewidth]{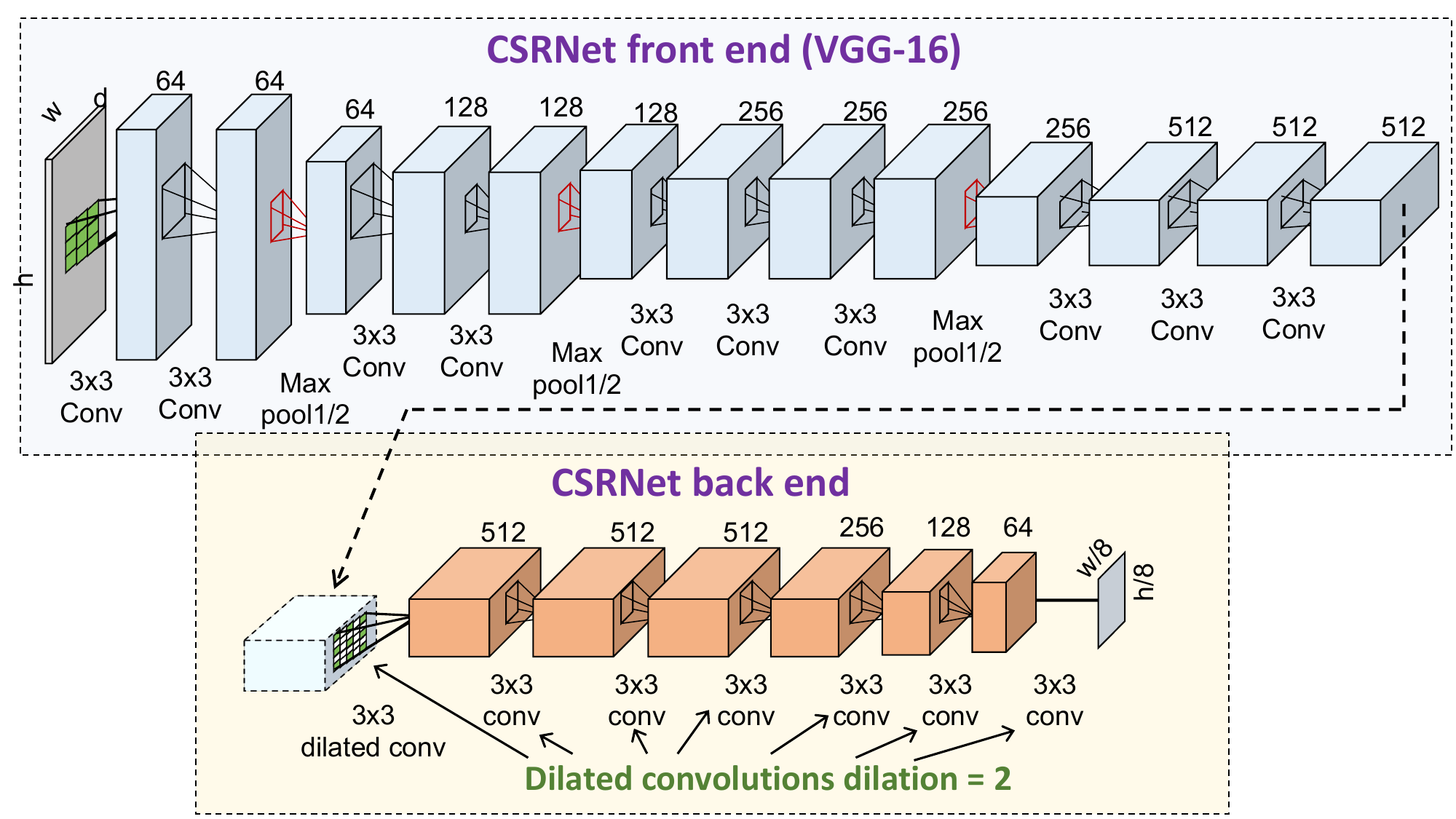}
  \end{center}
  \caption{The CSRNet architecture  
  }\label{fig:CSR_arch}
\end{figure*}

In this section we show the competing CNN architectures.  Figure~\ref{fig:ccnn_arch} shows the original Counting CNN (CCNN) network in \cite{Onoro:16}.  Figure~\ref{fig:mcnn_arch} shows the Multi-column CNN network (MCNN).  MCNN consists of 3 columns of CCNN-like structure where the convolutional kernel sizes are fixed at 3, 5 and 7 respectively.  Finally, Figure~\ref{fig:CSR_arch} shows CSRNet that consists of the first 10 layers of the VGG-16 model.  We refer to this as the front-end.  The back-end consists of 6 layers with dilated convolutional kernels to estimate the image density map.

\section{Superpixel Segmentation}
\label{sec:superpix_seg}

\begin{figure*}[h]
  \begin{center}
  \includegraphics[width=\linewidth]{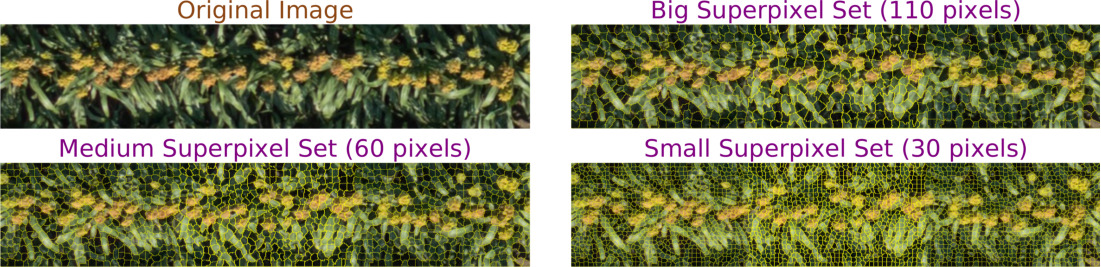}
  \end{center}
  \caption{An image and three levels of superpixel segmentation with
    average size of 30, 60 and 110 pixels.}\label{fig:slic}
\end{figure*}

Figure \ref{fig:slic} shows an example image along with three sets of superpixels used in the annotation tool.  We denoted the three sets small, medium and large superpixels in which the average superpixel size was respectively 30, 60 and 110 pixels.  The advantage of having these three segmentation sets is two-fold in both speed and accuracy of the annotation.  Firstly, it allows the annotator to save time by using the segmentation set whose superpixel sizes matches the panicle size most closely.  Secondly, having three independently made segmentation sets allow us to switch the segmentation set when the alignment to the actual panicles do not match the selected segmentation set.

% \subsection{Annotation Tool}

% Figure~\ref{fig:tool} shows the annotation tool in action.  The user applies the automatic instance segmentation on the active zoom level and can intermingle it with user annotations/corrections.

% \begin{figure*}[h]
% \begin{center}
%     \includemedia[
%       label=cloudy,
%     addresource=segmentation.mp4,
%     width=0.9\linewidth,height=0.30\linewidth,
%     activate=pageopen,
%     deactivate=onclick,
%     passcontext, %show VPlayer's rightclick menu
%     flashvars={
%       source=segmentation.mp4
%       &loop=true
%     },
%   ]{}{VPlayer.swf}
%   \mediabutton[
%     mediacommand=cloudy:playPause,
%     overface=\color{blue}{\fbox{\strut Play/Pause}},
%     downface=\color{red}{\fbox{\strut Play/Pause}}
%   ]{\fbox{\strut Play/Pause}}
%   \end{center}

%   \caption{The annotation tool in action, showing the instance segmentation in use.}\label{fig:tool}
% \end{figure*}

\section{Results From Best Model}
\label{sec:best_res}

\begin{figure*}[t]
  \begin{center}
  \includegraphics[width=\linewidth]{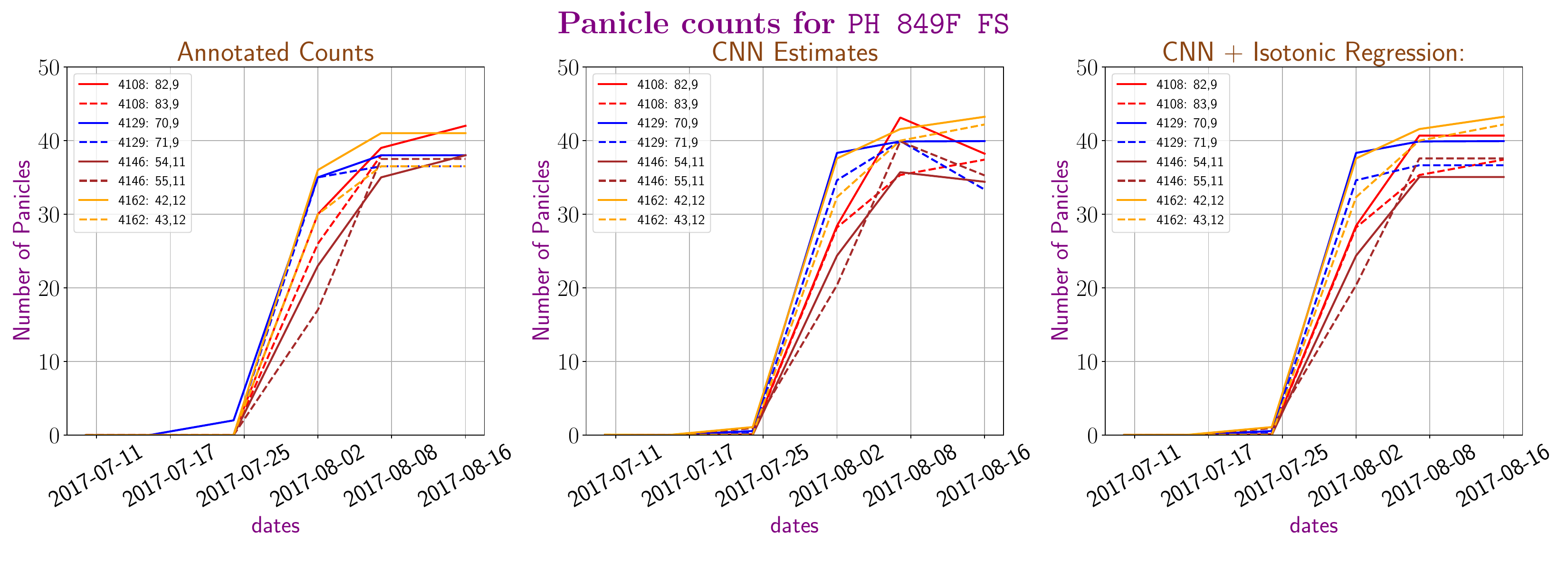}
  \includegraphics[width=\linewidth]{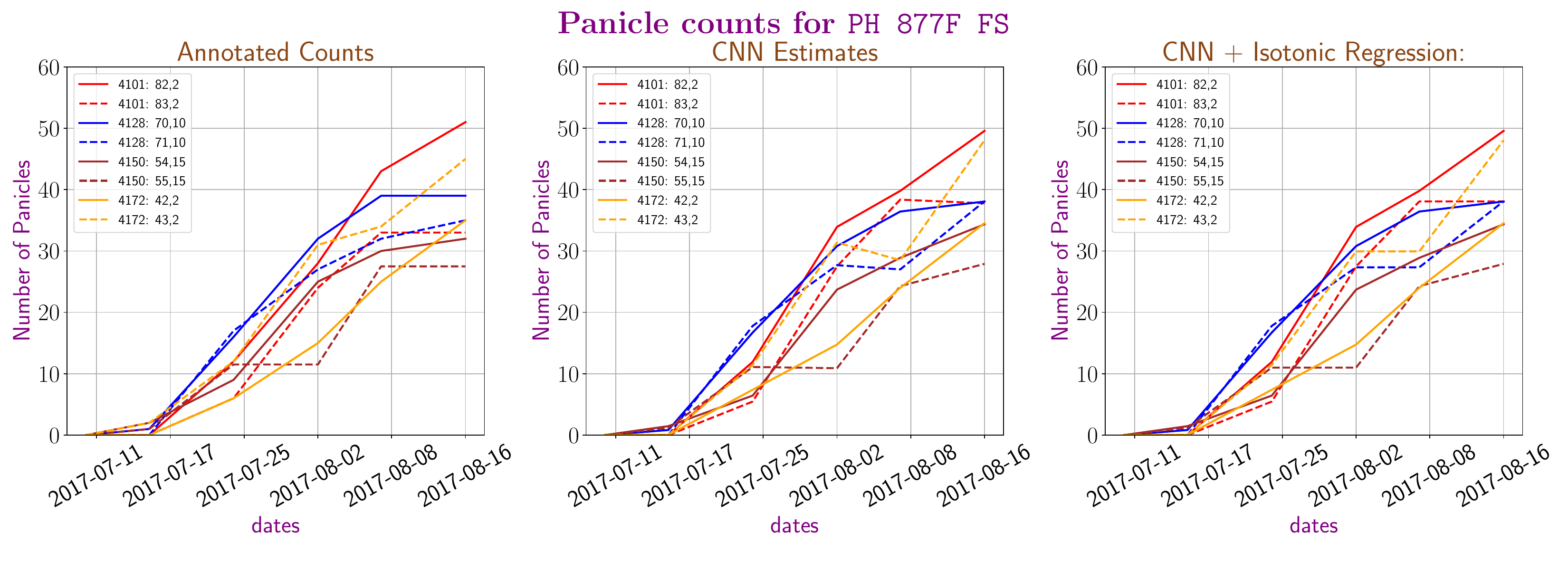}
  \includegraphics[width=\linewidth]{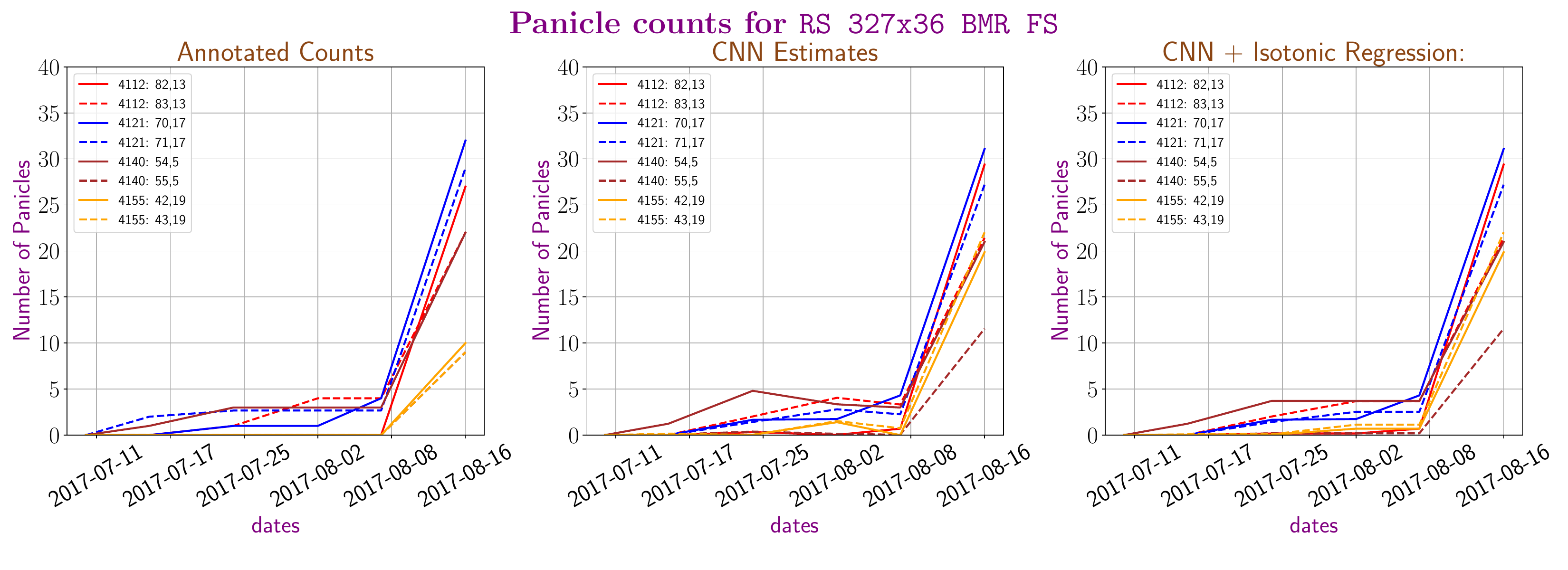}
  \end{center}
  \caption{Manual panicle counts and CNN prediction counts with and without isotonic regression for the best CNN system for the three varieties \texttt{PH 849F FS},
  \texttt{PH 877F FS} and
  \texttt{RS 327x36 BMR FS}.
  }\label{fig:isotonic1}
\end{figure*}

\begin{figure*}[t]
  \begin{center}
  \includegraphics[width=\linewidth]{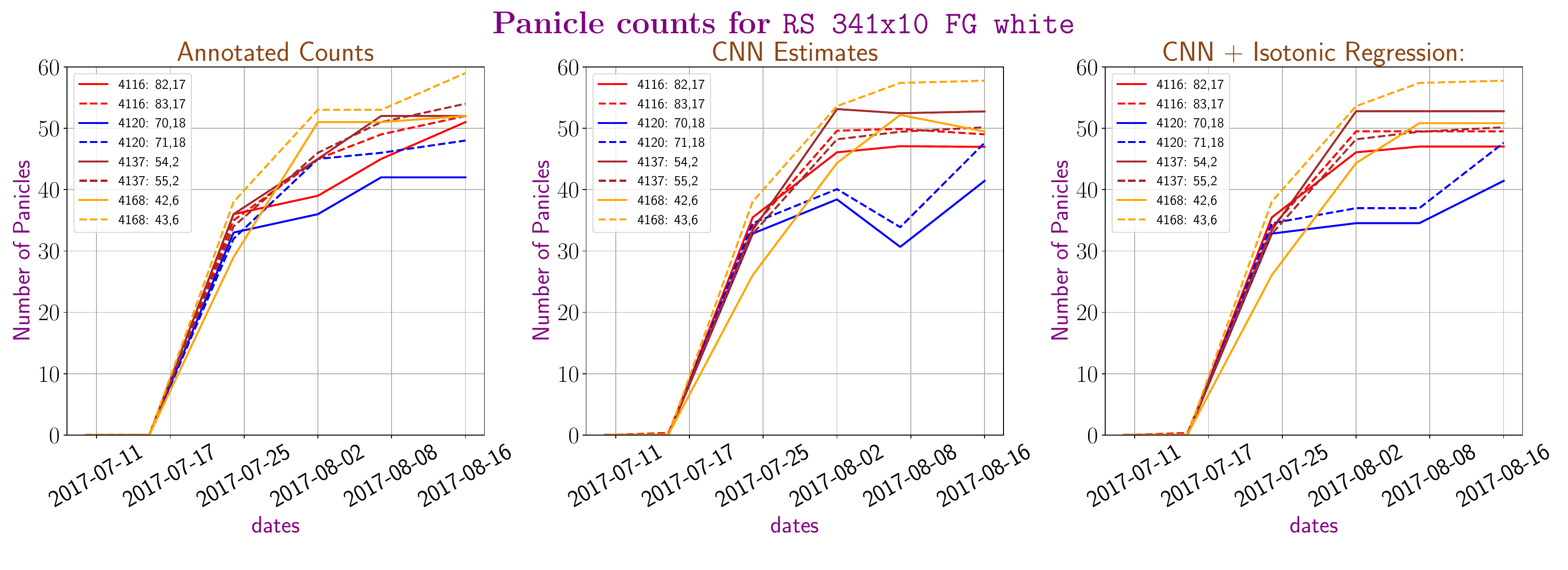}
  \includegraphics[width=\linewidth]{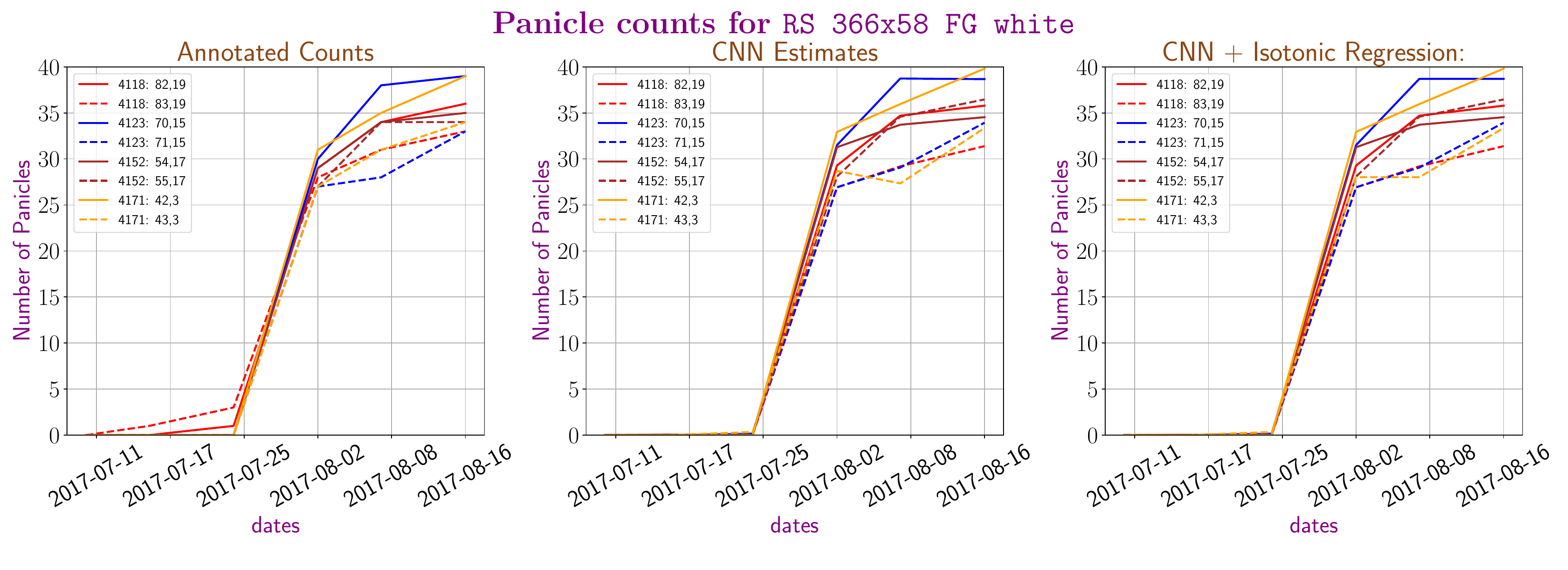}
  \includegraphics[width=\linewidth]{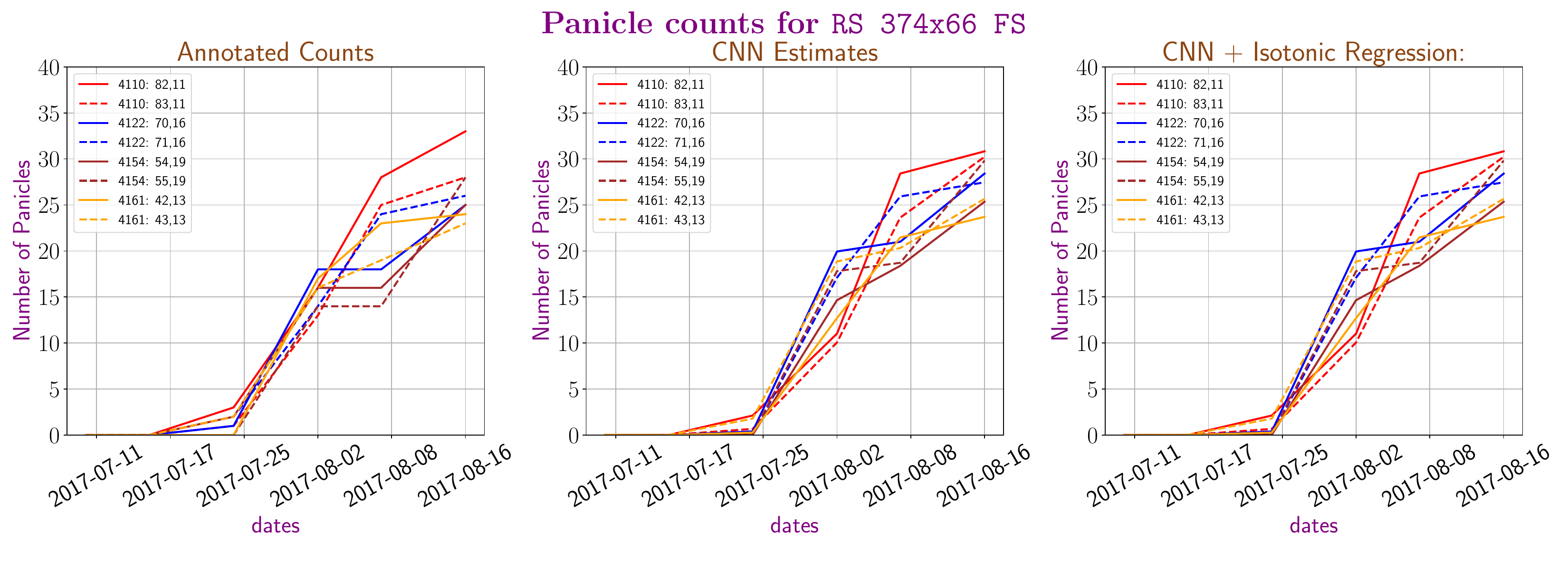}
  \end{center}
  \caption{Manual panicle counts and CNN prediction counts with and without isotonic regression for the best CNN system for the three varieties 
    \texttt{RS 341x10 FG white},
  \texttt{RS 366x58 FG white} and
  \texttt{RS 374x66 FS}.
  }\label{fig:isotonic2}
\end{figure*}

\begin{figure*}[t]
  \begin{center}
  \includegraphics[width=\linewidth]{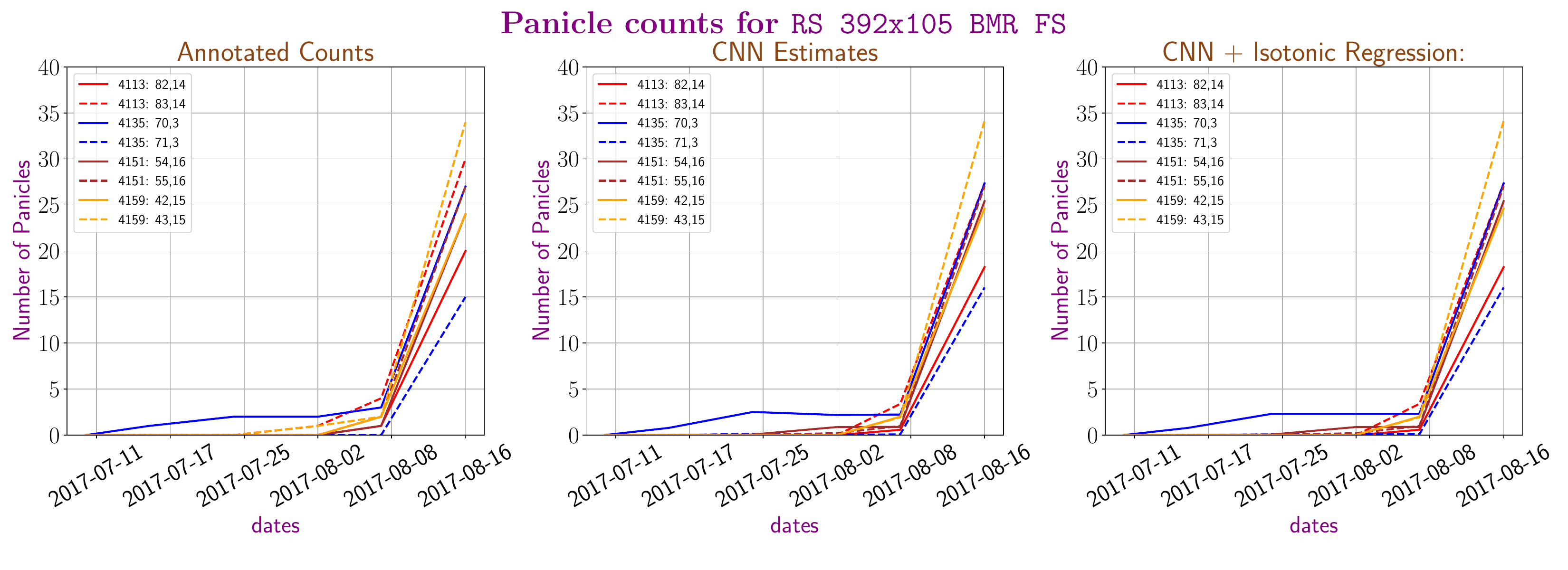}
  \includegraphics[width=\linewidth]{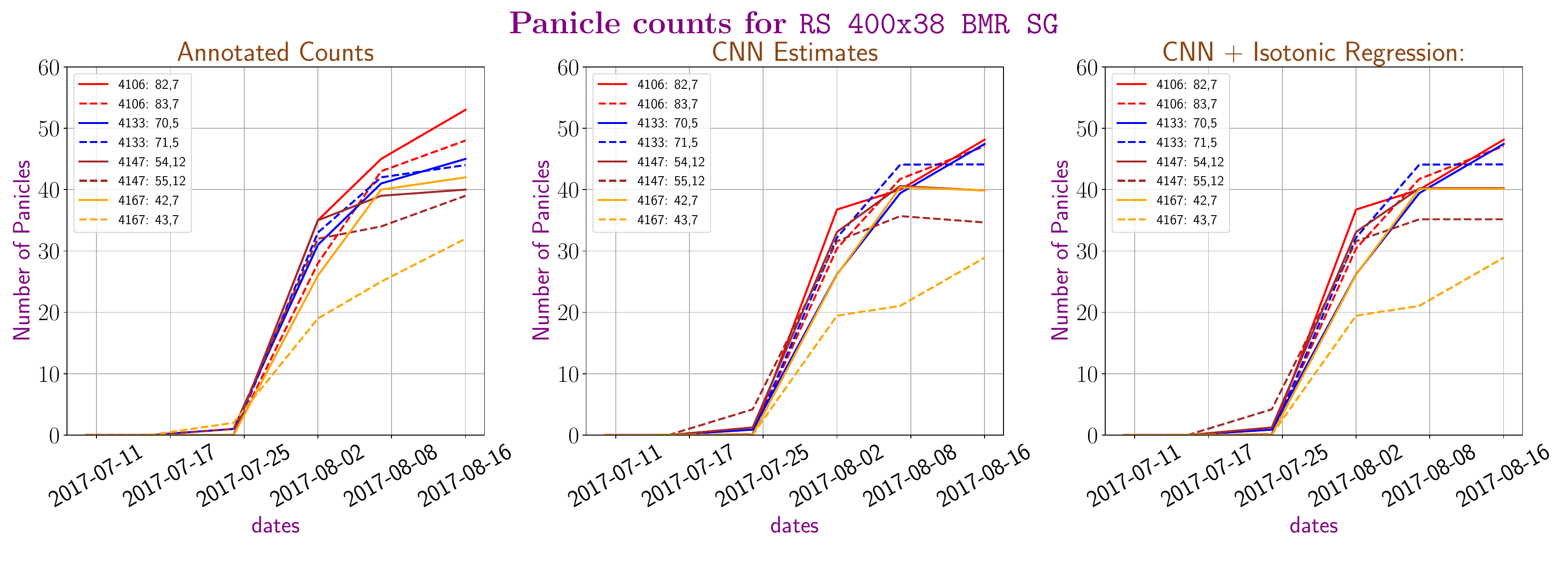}
  \includegraphics[width=\linewidth]{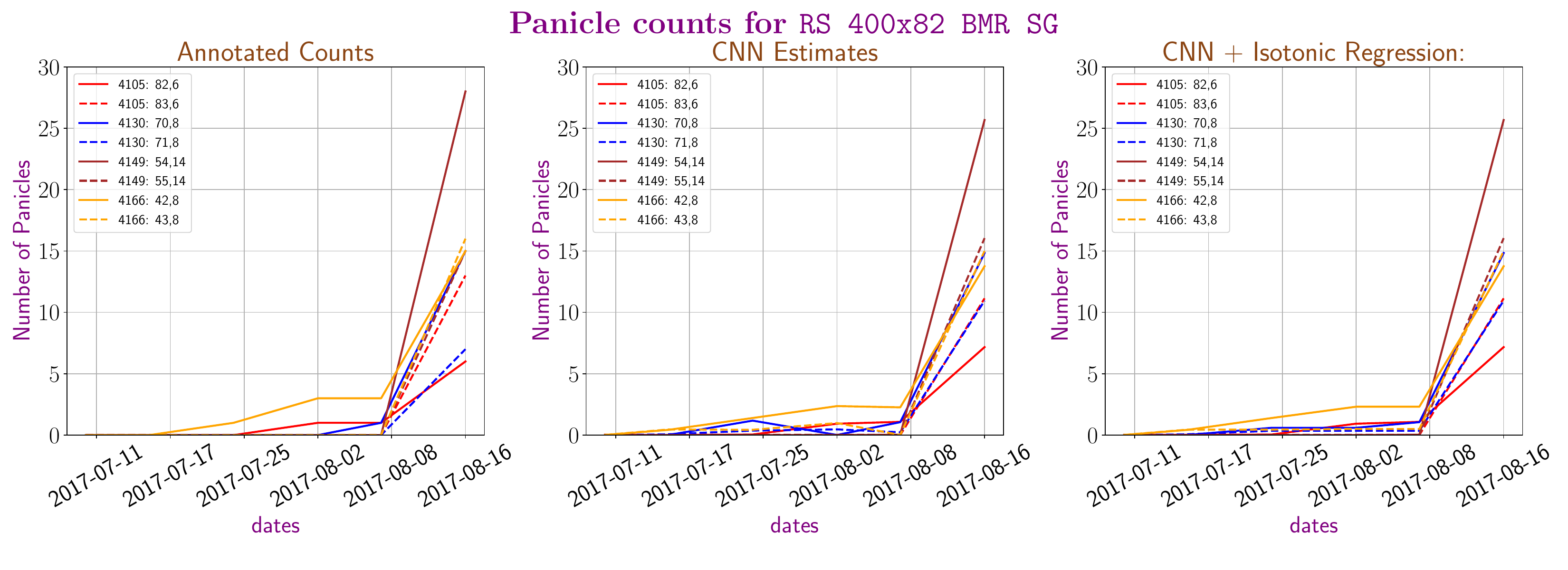}
  \end{center}
  \caption{Manual panicle counts and CNN prediction counts with and without isotonic regression for the best CNN system for the three varieties 
  \texttt{RS 392x105 BMR FS},
  \texttt{RS 400x38 BMR SG} and
  \texttt{RS 400x82 BMR SG}.
  }\label{fig:isotonic3}
\end{figure*}

\begin{figure*}[t]
  \begin{center}
  \includegraphics[width=\linewidth]{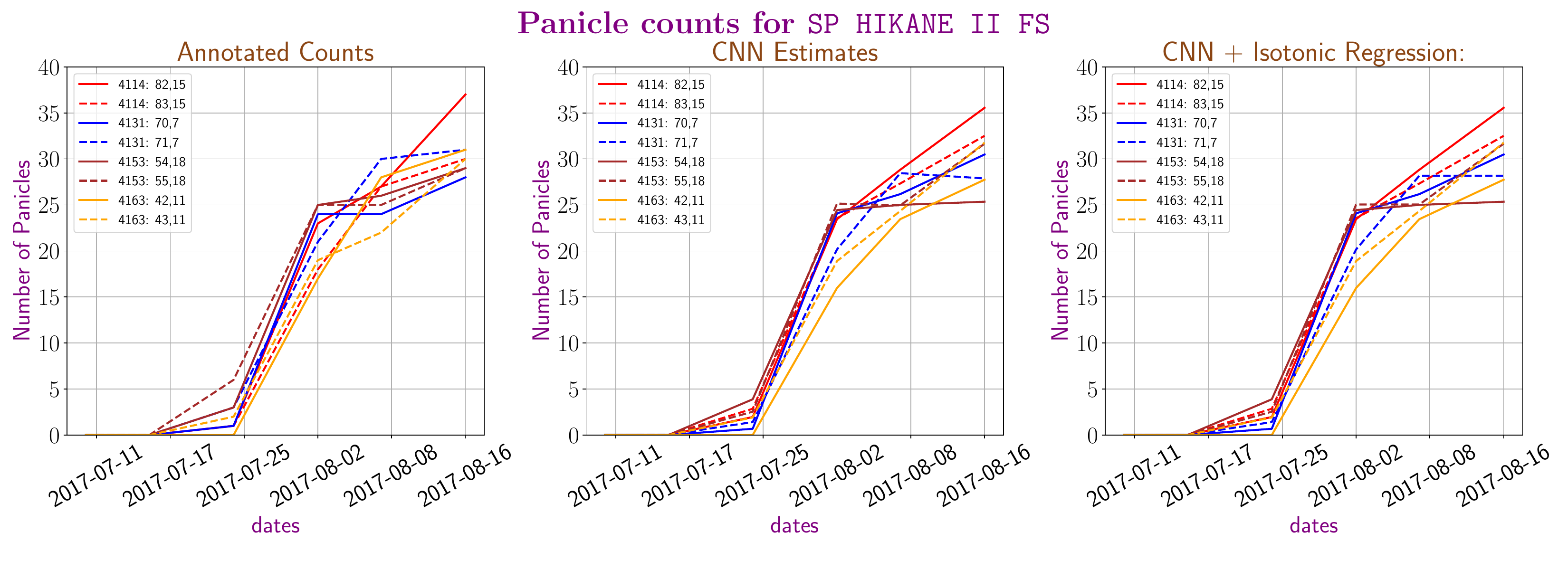}
  \includegraphics[width=\linewidth]{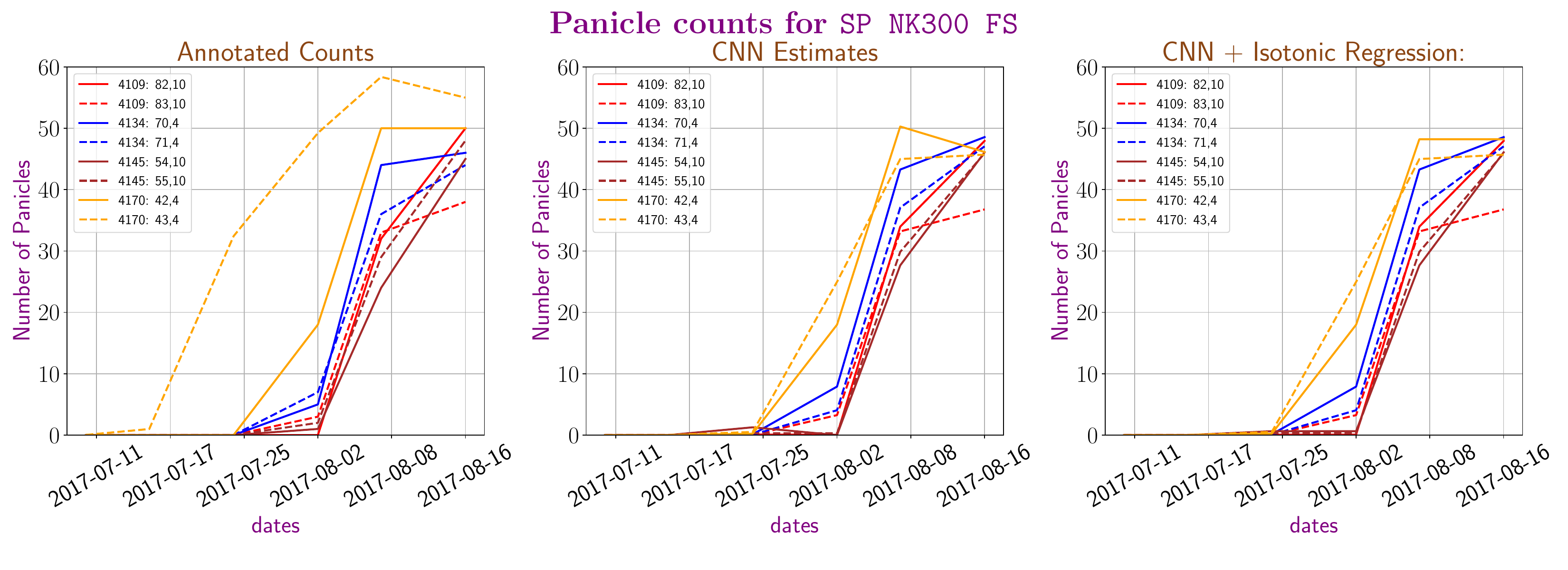}
  \includegraphics[width=\linewidth]{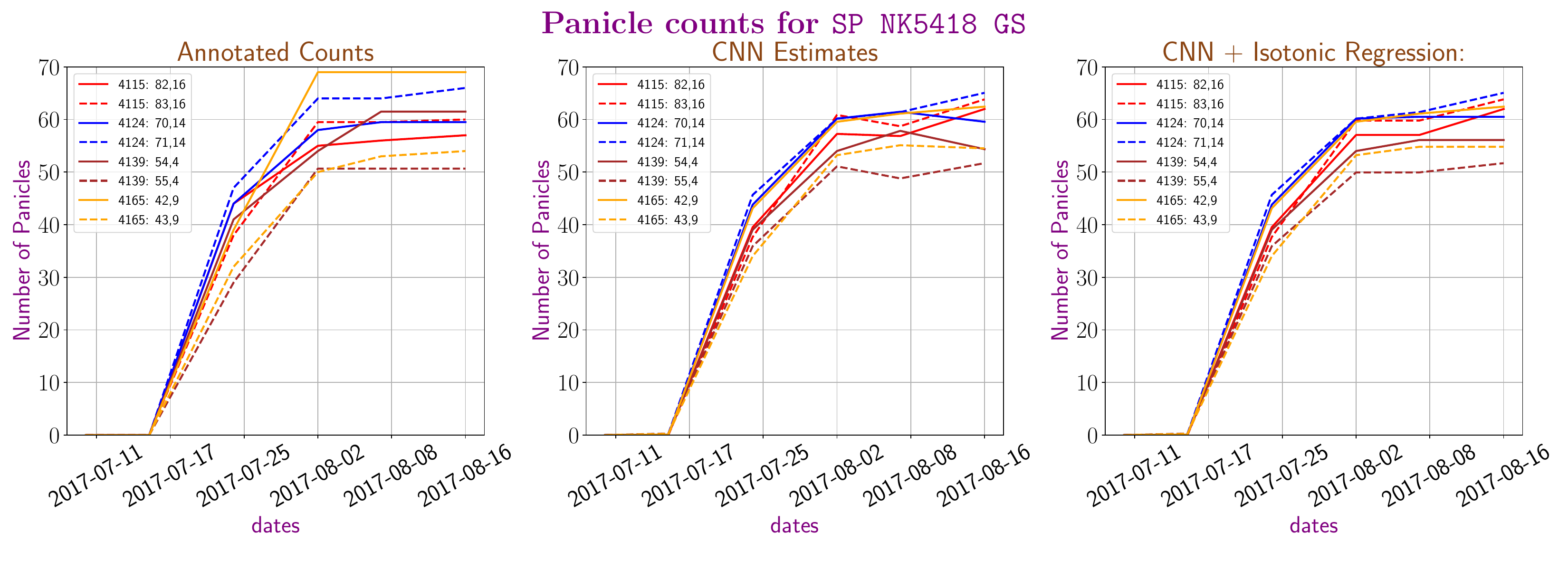}
  \end{center}
  \caption{Manual panicle counts and CNN prediction counts with and without isotonic regression for the best CNN system for the three varieties 
  \texttt{SP HIKANE II FS},
  \texttt{SP NK300 FS} and
 \texttt{SP NK5418 GS}.
  }\label{fig:isotonic4}
\end{figure*}

\begin{figure*}[t]
  \begin{center}
  \includegraphics[width=\linewidth]{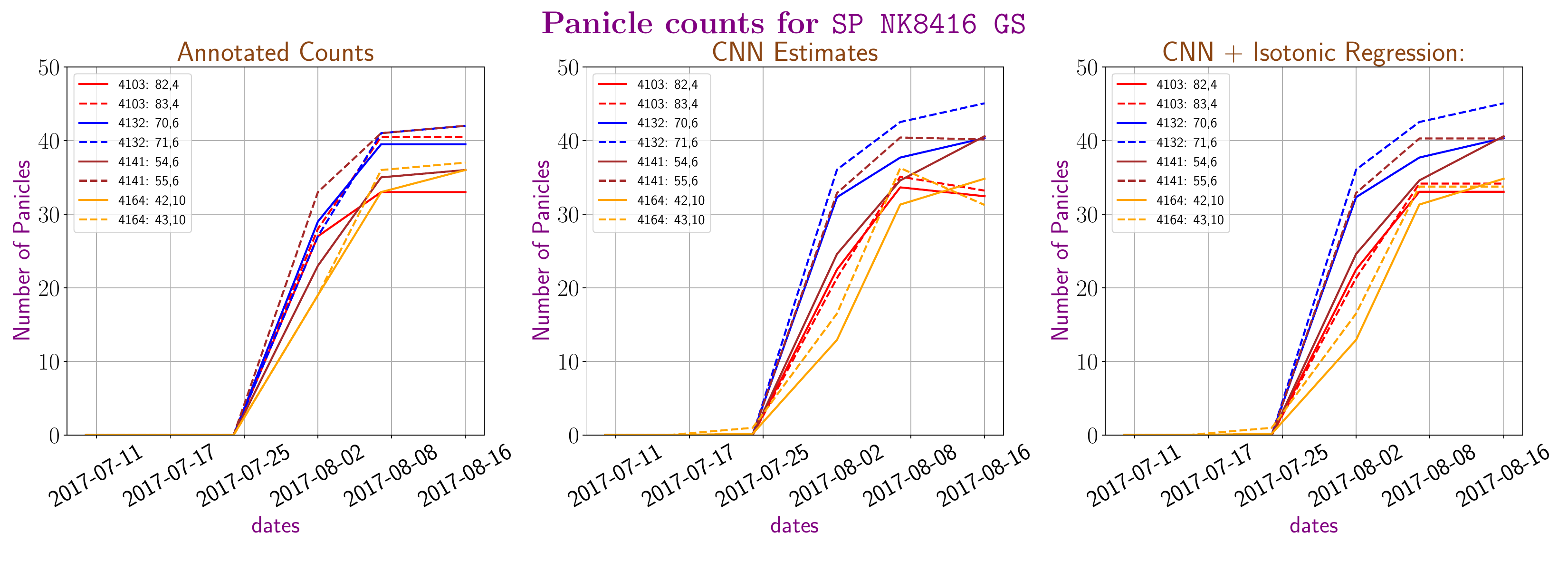}
  \includegraphics[width=\linewidth]{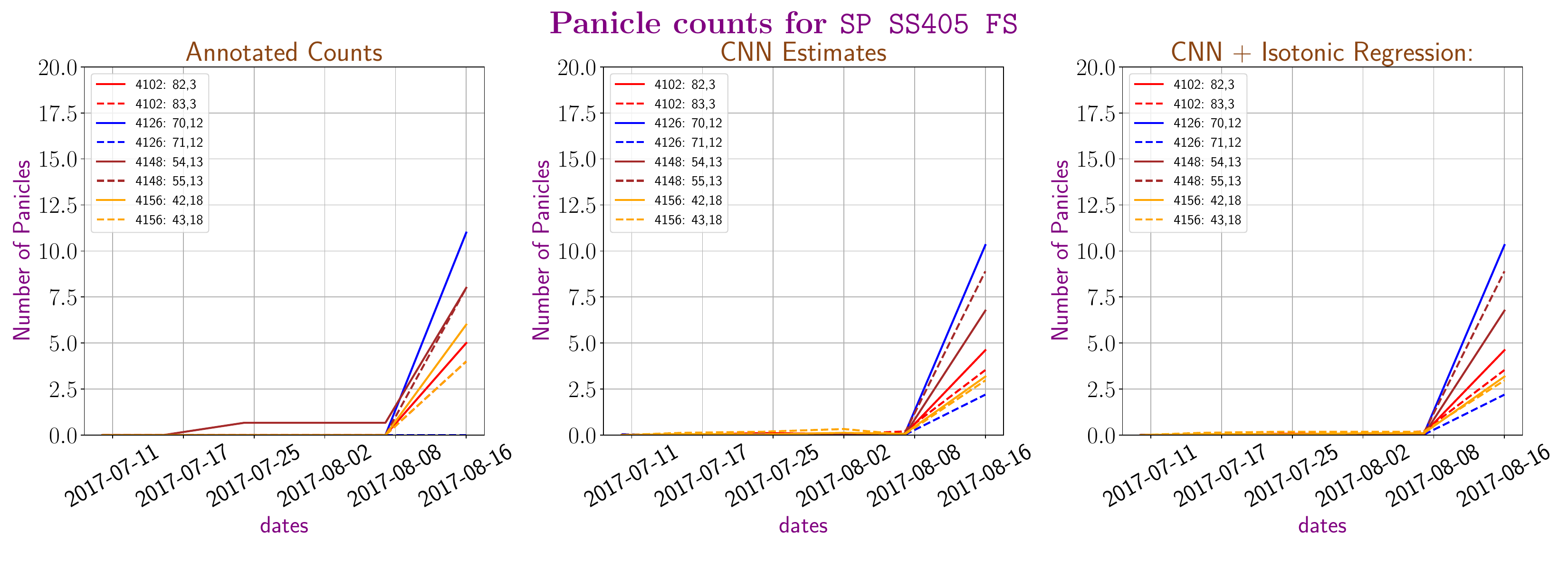}
  \includegraphics[width=\linewidth]{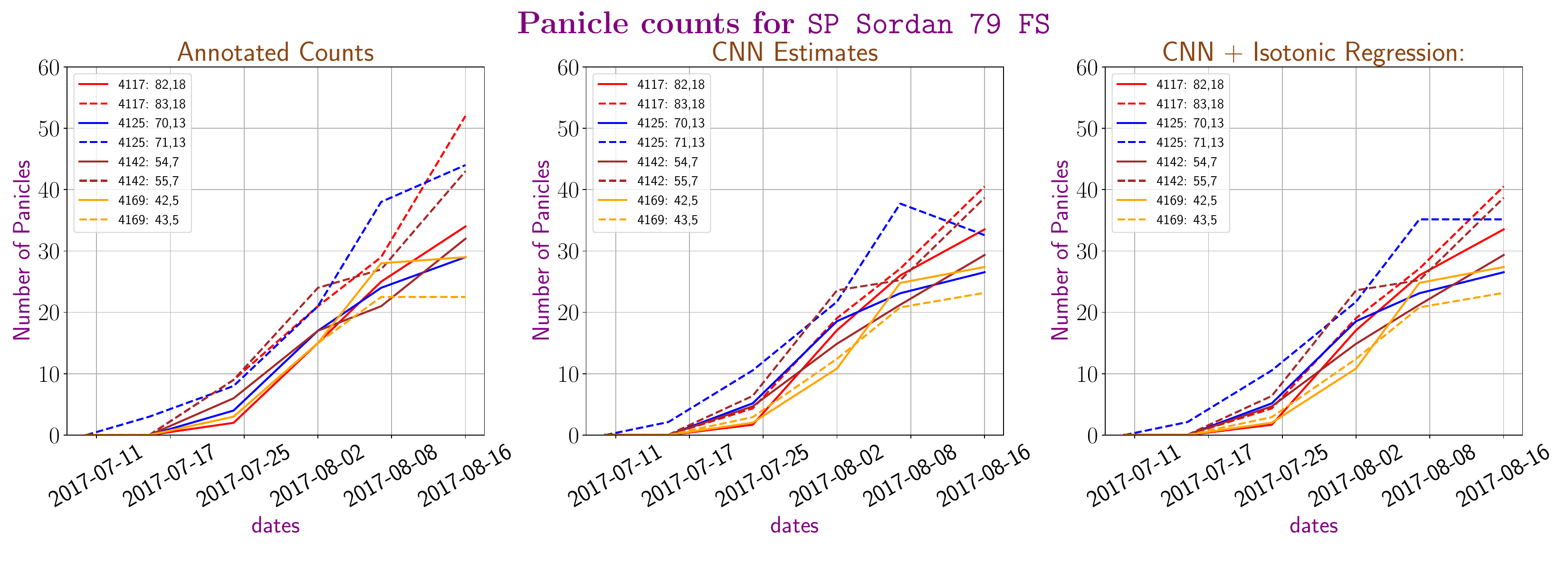}
  \end{center}
  \caption{Manual panicle counts and CNN prediction counts with and without isotonic regression for the best CNN system for the three varieties 
 \texttt{SP NK8416 GS},
 \texttt{SP SS405 FS} and
 \texttt{SP Sordan 79 FS}.
  }\label{fig:isotonic5}
\end{figure*}

\begin{figure*}[t]
  \begin{center}
  \includegraphics[width=\linewidth]{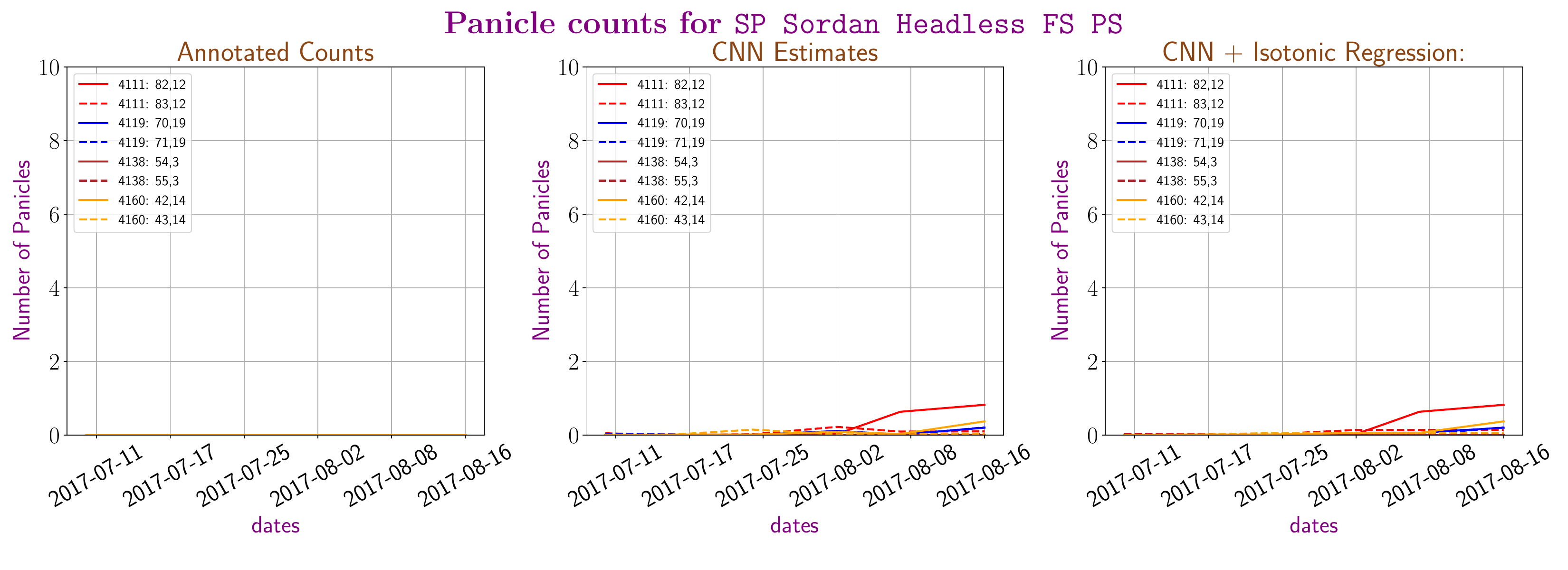}
  \includegraphics[width=\linewidth]{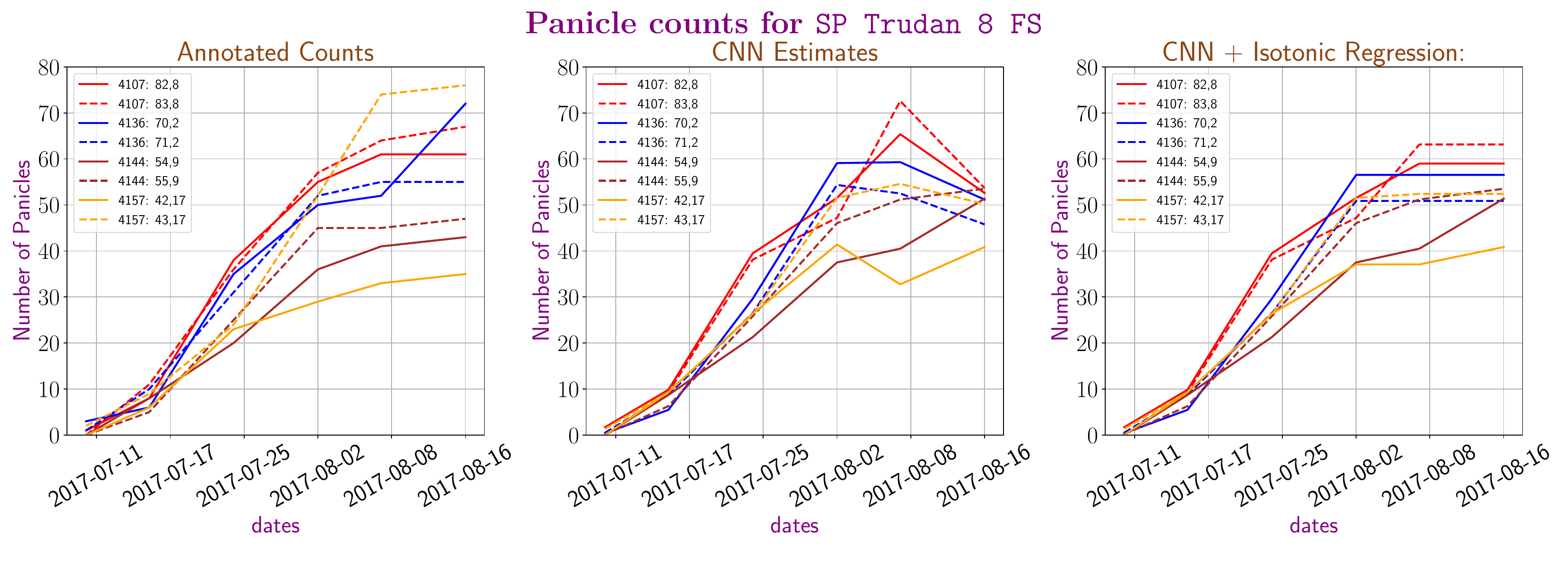}
  \includegraphics[width=\linewidth]{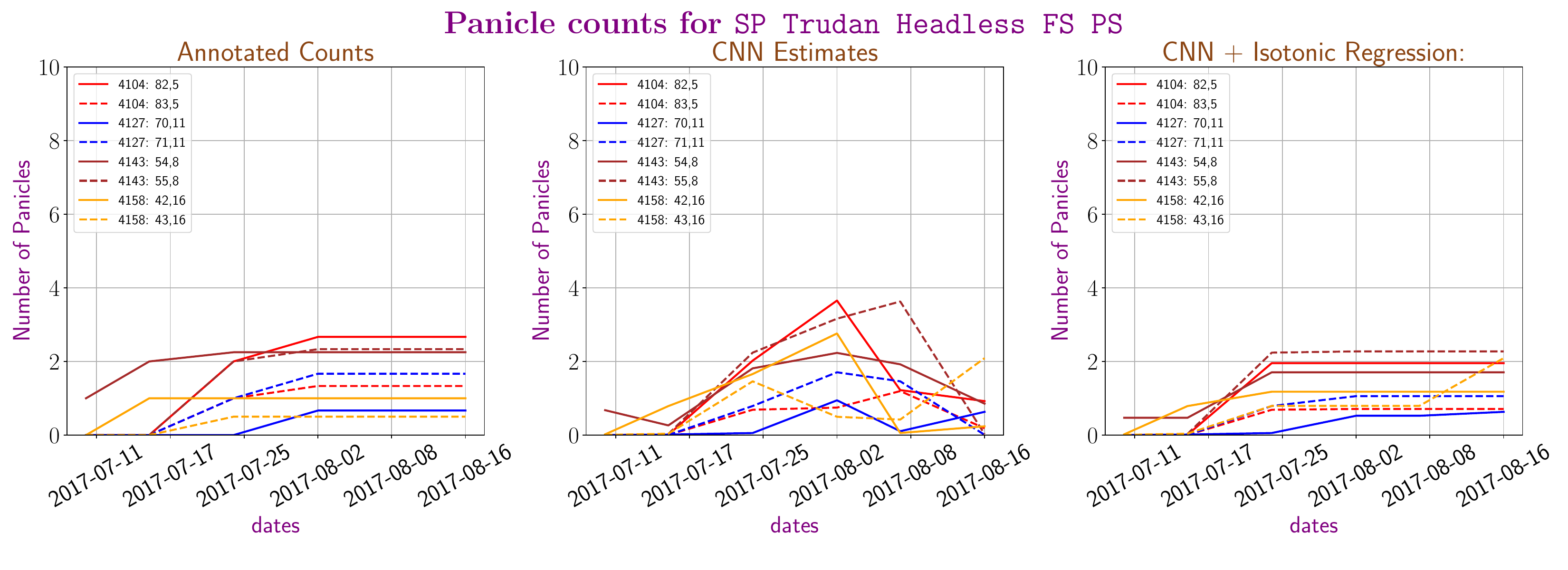}
  \end{center}
  \caption{Manual panicle counts and CNN prediction counts with and without isotonic regression for the best CNN system for the three varieties 
 \texttt{SP Sordan Headless FS PS},
 \texttt{SP Trudan 8 FS} and
 \texttt{SP Trudan Headless FS PS}. 
  }\label{fig:isotonic6}
\end{figure*}

The plots in Figure~\ref{fig:isotonic1}-\ref{fig:isotonic6} contains the counts and the predicted counts with and without isotonic regression for our best system (with MAE=1.17).  It can be seen that several of the varieties had very few panicles at all.  These varieties were photo-sensitive and only flowered when the days became shorter and we have no image from that part of the growing season.

\end{document}